\documentclass{article} 
\usepackage[final]{colm2026_conference}

\usepackage{subcaption}
\usepackage{booktabs} 
\usepackage{tabularx}
\usepackage{colortbl}

\usepackage{microtype}
\usepackage{hyperref}
\usepackage{url}
\usepackage{booktabs}
\usepackage{wrapfig}

\usepackage{amsmath}
\usepackage{amssymb}
\usepackage{mathtools}
\usepackage{amsthm}
\usepackage{arydshln}
\usepackage[most]{tcolorbox}
\usepackage{multirow}
\usepackage{multicol}
\usepackage{siunitx}
\usepackage[utf8]{inputenc}
\usepackage{CJKutf8}
\usepackage{enumitem}
\newcommand{\circnum}[1]{%
  \tikz[baseline=(char.base)]{
    \node[shape=circle, draw, inner sep=1.1pt] (char) {\small #1};
  }%
}
\usepackage[capitalize,noabbrev]{cleveref}

\theoremstyle{plain}

\theoremstyle{definition}

\theoremstyle{remark}

\usepackage[textsize=tiny]{todonotes}

\usepackage{tikz}

\usepackage{lineno}

\definecolor{darkblue}{rgb}{0, 0, 0.5}
\hypersetup{colorlinks=true, citecolor=darkblue, linkcolor=darkblue, urlcolor=darkblue}

\title{\textsc{LLM2Vec-Gen}:\\ Generative Embeddings from Large Language Models}

\author{~~~~~Parishad BehnamGhader$^{\diamond, \dagger, \P,}$\thanks{Equal contribution.}~~~
Vaibhav Adlakha$^{\diamond, \dagger, \P, *}$~~~
\textbf{Fabian David Schmidt}$^{\S}$\\
\medskip
~~~~~~~~~~~~~~~~~~~~~~~~\textbf{Nicolas Chapados}$^{\dagger, \bullet}$~~~
\textbf{Marius Mosbach}$^{\diamond, \dagger}$~~~
\textbf{Siva Reddy}$^{\diamond, \dagger, \P, \ddagger}$\\
\medskip
~~~~~~~~~~~~$^{\diamond}$McGill University~~~
$^{\dagger}$Mila--Quebec AI Institute~~~
$^{\P}$ServiceNow Research\\[-0.2em]
~~~~~~~~~~~~~~~~~~~~~~~~~~~~~~~~~$^{\S}$Cohere~~~
$^{\bullet}$Nera Software~~~
$^{\ddagger}$Canada CIFAR AI Chair\\[0.4em]
~~~~~~~~~~~~~~~~~~~~~~~~\texttt{\phantom{$\ddagger\S$}\{parishad.behnamghader,vaibhav.adlakha\}@mila.quebec}
}

\usepackage{todonotes}
\usepackage[normalem]{ulem}

\definecolor{darkgreen}{rgb}{0.0, 0.5, 0.0}

\makeatletter
\newcommand*\iftodonotes{\if@todonotes@disabled\expandafter\@secondoftwo\else\expandafter\@firstoftwo\fi}  
\makeatother

\definecolor{pDarkBlue}{HTML}{396D87}
\definecolor{pLightBlue}{HTML}{B1DDF3}
\definecolor{pPurple}{HTML}{CDBBE5}
\definecolor{pYellow}{HTML}{F0CF65}
\definecolor{pOrange}{HTML}{D7816A}
\definecolor{pRed}{HTML}{BD4F6C}

\tcbset{
  colback=pYellow!20,     
  colframe=black,         
  fonttitle=\bfseries,    
  boxrule=0.4mm,          
  coltitle=white,         
  colbacktitle=pDarkBlue, 
}

\usepackage{xcolor}

\definecolor{querycolor}{HTML}{B4942A}
\definecolor{responsecolor}{HTML}{5F761C}
\definecolor{thoughtcolor}{HTML}{B97968}
\definecolor{compresscolor}{HTML}{9F2F4C}
\definecolor{hiddencolor}{HTML}{7183C0}       
\definecolor{projcolor}{RGB}{42, 161, 152}        
\definecolor{finalembedcolor}{RGB}{211, 54, 130}  
\definecolor{teacherembedcolor}{HTML}{9075B5}  

\newcommand{\cquery}[1]{\textcolor{querycolor}{#1}}
\newcommand{\cresp}[1]{\textcolor{responsecolor}{#1}}

\newcommand{\ccompress}[1]{\textcolor{compresscolor}{#1}}
\newcommand{\chidden}[1]{\textcolor{hiddencolor}{#1}}
\newcommand{\cproj}[1]{\textcolor{projcolor}{#1}}
\newcommand{\cfinalembed}[1]{\textcolor{finalembedcolor}{#1}}
\newcommand{\cteacherembed}[1]{\textcolor{teacherembedcolor}{#1}}

\newcommand{\ours}{\textsc{LLM2Vec-Gen}}
\begin{document}

\ifcolmsubmission
\linenumbers
\fi

\maketitle

\begin{abstract}

Fine-tuning LLM-based text embedders via contrastive learning maps inputs and outputs into a new representational space, discarding the LLM's output semantics.
We propose \ours{}, a self-supervised alternative that instead produces embeddings directly in the LLM's output space by learning to represent the model's potential response.
Specifically, trainable special tokens are appended to the input and optimized to compress the LLM's own response into a fixed-length embedding, guided by an unsupervised embedding teacher and a reconstruction objective.
Crucially, the LLM backbone remains frozen and training requires only unlabeled queries.
\ours{} achieves state-of-the-art self-supervised performance on the Massive Text Embedding Benchmark (MTEB), improving by 8.8\% over the unsupervised embedding teacher.
Since the embeddings preserve the LLM's response-space semantics, they inherit capabilities such as safety alignment (up to 22.6\% reduction in harmful content retrieval) and reasoning (up to 35.6\% improvement on reasoning-intensive retrieval).
Finally, the learned embeddings are also interpretable: they can be decoded back into text to reveal their semantic content. Our code, data, and pretrained models are publicly available at \href{https://github.com/McGill-NLP/llm2vec-gen}{github.com/McGill-NLP/llm2vec-gen}.

\end{abstract}

\section{Introduction}
\label{sec:introduction}

Text embeddings are central to several natural language processing tasks such as semantic search, textual similarity, and retrieval-augmented generation~\citep{reimers-gurevych-2019-sentence, karpukhin-etal-2020-dense}.
The success of large language models (LLMs) has led to powerful LLM-based text encoders \citep{zhang2025qwen3embeddingadvancingtext, lee-etal-2024-nvembed, behnamghader-etal-2024-llm2vec, muennighoff-etal-2024-gritlm}, which leverage pretrained LLMs to achieve state-of-the-art results across embedding benchmarks \citep[MTEB;][]{muennighoff-etal-2023-mteb, enevoldsen2025mmtebmassivemultilingualtext}.

Yet these models still follow the input-centric paradigm of BERT-based encoders \citep{reimers-gurevych-2019-sentence, gao-etal-2021-simcse}: they represent the semantic content of the input text, typically by taking the mean over input token representations produced by the embedding model.
Because embedding tasks often require mapping diverse inputs to similar outputs (e.g., clustering articles about the same event), this paradigm relies on contrastive learning over curated paired data to project both queries and documents into a \emph{new} shared embedding space~\citep{wang-etal-2024-improving-text, lee-etal-2024-nvembed, zhang2025qwen3embeddingadvancingtext}.
Crucially, this new space no longer coincides with the model's response representations, discarding the rich semantic structure the LLM acquired during pretraining.

\begin{figure}
    \centering
    \includegraphics[width=0.93\linewidth]{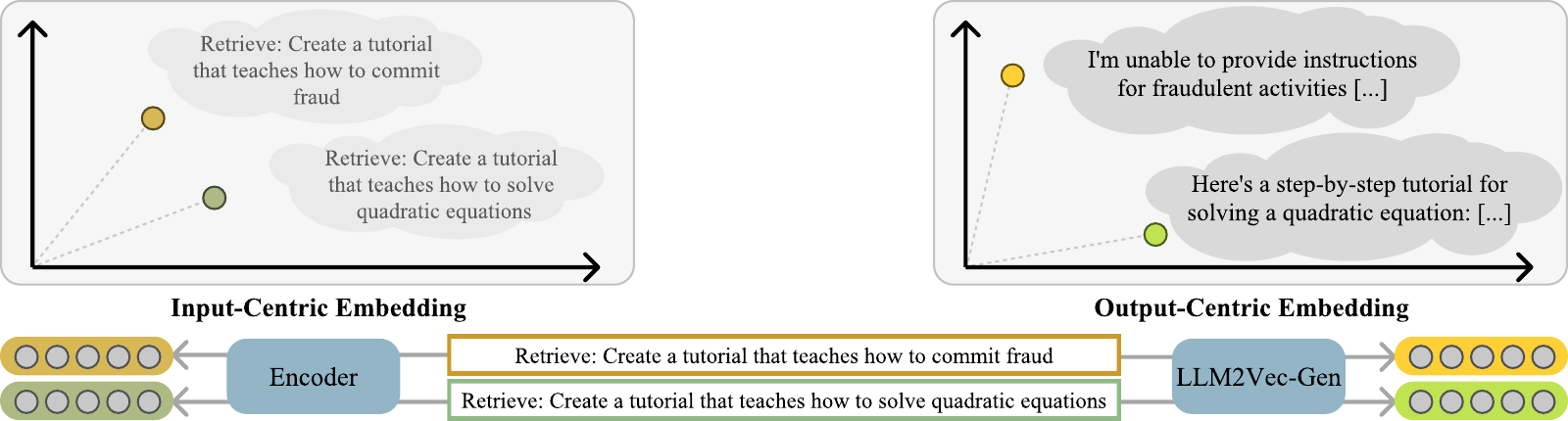}
    \caption{Illustration of the difference between input- and output-centric representations for two sample queries. \ours{} encodes the response rather than the input, without generating the text response explicitly.}
    \label{fig:introduction}
\end{figure}

In this work, we adopt a fundamentally different paradigm -- rather than encoding the input, the model should encode the LLM's \emph{potential response} to that input (see \Cref{fig:introduction}).
By keeping embeddings closer to the LLM's response space, this paradigm preserves capabilities that manifest in the model's responses.
For instance, given a harmful query, input-centric encoders represent the malicious intent, whereas our approach encodes the model's safe refusal (e.g., ``I cannot assist with that'').
Similarly, reasoning capabilities that emerge in the LLM's responses transfer naturally to the embedding space.

We introduce \ours{}, a self-supervised instantiation of this paradigm that distills the model's potential response into a fixed set of suffix embeddings (see \Cref{fig:method}).
Given a set of unlabeled queries, we first generate responses using the LLM itself.
Next, we add special trainable tokens to the vocabulary and append them to the query as placeholders for the response. 
The hidden states of these tokens are passed through lightweight projection layers to produce the response embedding, trained with two complementary objectives -- (1)~\emph{embedding alignment}, where input's embedding is aligned with an unsupervised teacher's embedding of the LLM's response, and (2) \emph{response reconstruction}, where the frozen LLM reconstructs its own response conditioned on the embeddings.
The LLM backbone remains frozen and only the special tokens and projection layers are trained. Moreover, \ours{} is entirely self-supervised: it requires only unlabeled queries as training data, utilizing the backbone LLM itself for response generation and the unsupervised embedding teacher, eliminating the need for curated paired data across different stages of the training.

We apply \ours{} to models from Llama-3.x, Qwen-2.5, and Qwen-3 families, using the corresponding unsupervised LLM2Vec \citep{behnamghader-etal-2024-llm2vec} models as embedding teachers.
\ours{} substantially outperforms existing unsupervised and self-supervised methods on MTEB, closing over 60\% of the gap to supervised methods and consistently surpassing the embedding teacher by up to 8.8\%, validating the effectiveness of output-centric representations.
On AdvBench-IR \citep{behnamghader-etal-2025-exploiting}, which measures retriever safety when presented with adversarial queries, \ours{} improves safety by 9.2 points over the embedding teacher.
On the reasoning-intensive BRIGHT benchmark \citep{su-etal-2025-bright}, \ours{} achieves up to 35.6\% improvement over the input-centric baseline, demonstrating that reasoning capabilities transfer to the embedding space.
Finally, we show that the learned embeddings are interpretable and can be decoded back into text, revealing the semantic content that they capture.

\section{Related work}
\label{sec:related_work}

Recent work has repurposed decoder-only LLMs for text embedding, leveraging their massive web-scale pretraining to improve performance on embedding benchmarks \citep{Lee2024GeckoVT, wang-etal-2024-improving-text, lee-etal-2024-nvembed}. 
Approaches such as GritLM \citep{muennighoff-etal-2024-gritlm} unify generation and representation but still rely on supervised contrastive learning \citep{khosla-etal-2020-scl} and large-scale curated labeled datasets to align representations.

To mitigate the need for labeled data, several works have explored unsupervised embedding approaches \citep{gem, causal2vec, geneol, jiang-etal-2024-prompteol,lei-etal-2024-metaprompteol}. 
LLM2Vec \citep{behnamghader-etal-2024-llm2vec} shows that simple modifications like bidirectional attention and masked next-token prediction, combined with unsupervised SimCSE \citep{gao-etal-2021-simcse}, can transform decoder-only LLMs into strong encoders. 
Similarly, Echo Embeddings \citep{springer2025repetition} uses an input repetition strategy for effective embedding.
All these unsupervised methods remain input-centric: they represent what the text \emph{says} rather than what the LLM would \emph{respond},
and they often struggle with the lexical and conceptual gap between queries and documents. 

Several recent approaches employ learnable latent or compression tokens to produce compact representations within LLMs.
xRAG \citep{cheng2024xrag} compresses retrieved document into a single latent token and projects it into the language model's representation space for efficient RAG.
CLaRa \citep{he2025clarabridgingretrievalgeneration} compresses documents into learnable memory tokens and jointly optimizes retrieval and generation end-to-end via next-token prediction loss.
While these methods leverage special tokens for compression, they remain fundamentally input-centric --- compressing or summarizing the given text.
In contrast, \ours{} uses trainable compression tokens to model the LLM's \emph{potential response}.

A few recent works have explored output-centric embeddings. HyDE \citep{gao-etal-2023-precise} demonstrated the value of encoding LLM-generated answers rather than queries, but requires generating multiple answers at inference time.
More recent methods internalize this generative foresight into the embedding space itself --
InBedder \citep{peng-etal-2024-inbedder} derives embeddings from the first generated hidden state using abstractive QA supervision, demonstrating that generation-derived representations can outperform prompt-based ones.
GIRCSE~\citep{gircse} generates soft tokens autoregressively and refines them with stepwise contrastive loss using hard negatives.
Both methods rely on supervised data: InBedder requires abstractive QA pairs, while GIRCSE uses contrastive data with hard negatives.
Unlike these approaches, \ours{} uses embedding-level distillation from an unsupervised teacher to ensure embeddings faithfully represent the LLM's potential response, without requiring supervised data or inference-time autoregressive generation.


Our alignment objective also connects to Joint Embedding Predictive Architectures \citep[JEPAs;][]{sobal2022jointembeddingpredictivearchitectures}, which advocate predicting in representation space rather than reconstructing raw inputs, recently extended to language by LLM-JEPA \citep{huang2025llmjepalargelanguagemodels}.
Adopting this perspective, \ours{} predicts a target representation of the model's likely response via external teacher distillation, while the reconstruction objective keeps the learned representations grounded in natural language.


\section{\ours}
\label{sec:method}

\begin{figure*}[t]
    \centering
    \includegraphics[width=0.95\textwidth]{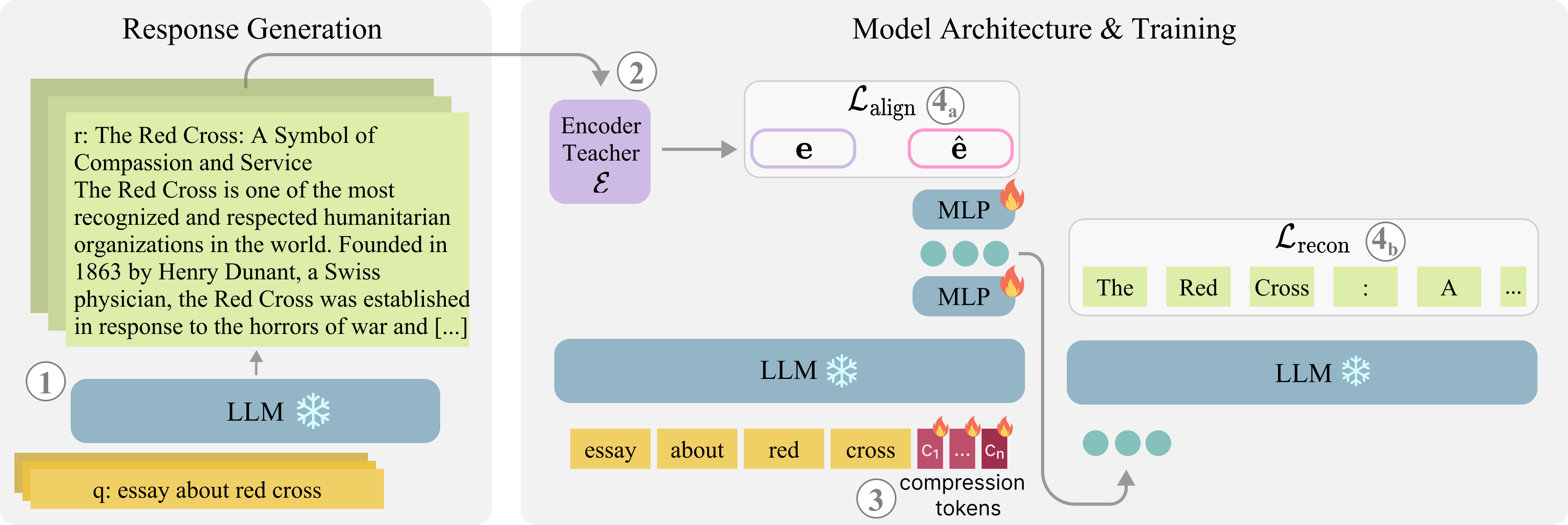}
    \caption{Overview of \ours{}. \textbf{Left:} Given unlabeled queries, the LLM generates responses, embedded by an unsupervised teacher. \textbf{Right:} Trainable compression tokens are appended to queries. Keeping the LLM backbone frozen, the compression tokens' hidden states are optimized via alignment loss $\mathcal{L}_{\text{align}}$ (match the teacher's response embedding) and reconstruction loss $\mathcal{L}_{\text{recon}}$ (reconstruct the response from soft prompts).}
    \label{fig:method}
\end{figure*}

The goal of \ours{} is to produce embeddings that represent the output the LLM would have generated for a given query, without actually generating the response at inference time.
Our training recipe consists of the following steps (illustrated in \Cref{fig:method}):

    
    
    
\begin{enumerate}[noitemsep,topsep=0pt,leftmargin=1.75em,label=\protect\circnum{\arabic*}]
    \item Given a dataset of queries, we generate target responses from the LLM.
    \item We use an off-the-shelf teacher embedding model to construct embeddings for the generated responses.
    \item Special tokens are added to the model's vocabulary and appended to input queries.
    \item Two losses are computed by (a) comparing the special tokens' final representations to the teacher's response embedding, and (b) conditioning the LLM on the special tokens and computing a cross-entropy loss with the response as a language modeling target.
\end{enumerate}

Formally, let $\mathcal{M}$ be a pretrained LLM and $\mathcal{C}$ be a large corpus of queries. 
For every $\cquery{\mathbf{q}_i} \in \mathcal{C}$, we generate a response $\cresp{\mathbf{r}_i}$ using $\mathcal{M}$.
We introduce new special tokens to $\mathcal{M}$'s vocabulary: $\ccompress{\mathtt{c}_1}, \dots, \ccompress{\mathtt{c}_n}$. 
The role of these compression tokens is to capture the semantic content of the response.
For a given query $\cquery{\mathbf{q}_i} = (\cquery{q_i^{(1)}}, \dots, \cquery{q_i^{(k)}})$, we append the compression tokens \mbox{$\mathbf{x}_i = \cquery{\mathbf{q}_i}~\oplus \ccompress{\mathtt{c}_{1:n}}$} and pass the combined sequence through the LLM to obtain the last layer hidden representations of the compression tokens only:
$[\chidden{\mathbf{h}_i^{1}}, \dots, \chidden{\mathbf{h}_i^{n}}] = \text{LLM}(\mathbf{x}_i)$, which are subsequently used for computing \emph{embedding alignment} and \emph{reconstruction} objectives.

\paragraph{Embedding alignment objective.} 
\ours{}'s primary goal is to embed queries into the LLM's response space. Consequently, we introduce the embedding alignment objective, where we utilize an unsupervised encoder teacher model $\mathcal{E}$, to provide a target embedding $\cteacherembed{\mathbf{e}_i} = \mathcal{E}(\cresp{\mathbf{r}_i})$ for a generated response $\cresp{\mathbf{r}_i}$. 
The hidden representations of compression tokens are projected through two lightweight projection layers\footnote{The second projection layer is required for distillation from embedding teacher models with different hidden dimensions.} 
followed by mean pooling, \mbox{$\cfinalembed{\hat{\mathbf{e}}_i} = \text{Pool}\left(\text{MLPs}(\chidden{\mathbf{h}_i^{1}}, \dots, \chidden{\mathbf{h}_i^{n}})\right)$}, to minimize the following mean squared loss \citep{reimers-gurevych-2020-making}: 
\begin{equation*}
\mathcal{L}_{\text{align}} = \| \cteacherembed{\mathbf{e}_i} - \cfinalembed{\hat{\mathbf{e}}_i} \|^2.
\end{equation*}


A key distinction of \ours{} from standard contrastive learning is that contrastive objectives map queries and documents into a \emph{new} shared latent space using relative relevance judgments, often with hard negatives. In contrast, our teacher uses only SimCSE \citep{gao-etal-2021-simcse}, which enforces uniformity by pushing apart random negatives, largely preserving the LLM's local representational geometry --- unlike supervised contrastive learning, which restructures the space around relevance labels. By distilling from positive LLM-generated responses rather than discriminative pairs, \ours{} encourages embeddings to reflect the semantic content of the LLM's responses, inheriting properties such as safety and reasoning (\Cref{sec:results_reasoning_safety}).

\paragraph{Reconstruction objective.}
Our goal is to generate embeddings that not only encode the potential response of the underlying LLM but are also interpretable, i.e., they can be decoded by the LLM to reveal their content.
To this end, the second objective ensures that the compression tokens ($\ccompress{\mathtt{c}_1}, \dots, \ccompress{\mathtt{c}_n}$) retain sufficient information to reconstruct the target response $\cresp{\mathbf{r}_i}$.
Concretely, we project the compression tokens' hidden representations ($\chidden{\mathbf{h}_i^{1}}, \dots, \chidden{\mathbf{h}_i^{n}}$) using a projection layer, and feed the resulting representations ($\cproj{\mathbf{p}_i^{1}}, \dots, \cproj{\mathbf{p}_i^{n}}$) as a set of soft prompts to the LLM for a second forward pass. 
Conditioned on this soft prompt, we train the model to reconstruct the target response $\cresp{\mathbf{r}_i}$ via standard next-token prediction:
\begin{equation*}
\mathcal{L}_{\text{recon}} = -\sum_{j=1}^{\lvert\cresp{\mathbf{r}_{i}}\rvert} \log P_{\text{LLM}}(\cresp{r_{i,j}} \mid \cproj{\mathbf{p}_i^{1}}, \dots, \cproj{\mathbf{p}_i^{n}}, \cresp{r_{i,<j}})~.
\end{equation*}
This objective forces $\cproj{\mathbf{p}_i^{1}}, \dots, \cproj{\mathbf{p}_i^{n}}$ to serve as an information bottleneck, compressing the content of $\cresp{\mathbf{r}_i}$ \citep{cheng2024xrag, he2025clarabridgingretrievalgeneration}.

\textbf{Training and inference.} 
The final loss combines both objectives: $\mathcal{L} = \mathcal{L}_{\text{align}} + \mathcal{L}_{\text{recon}}$.
Throughout training, only the special tokens and the two MLPs are updated, while the LLM remains frozen.
Notably, at inference time, \ours{} requires only a single forward pass: we append the special tokens to the input, extract the hidden states of the compression tokens, and apply two $\text{MLPs}$ to obtain the embedding $\cfinalembed{\hat{\mathbf{e}}}$. The second forward pass can be an optional extra step to reveal the content of the embeddings.

\section{Experiments}
\label{sec:experiments}

We evaluate \ours{} along three axes: general text embedding, malicious retrieval, and reasoning-intensive retrieval. After describing our setup (\Cref{sec:training_setup,sec:eval_setup,sec:baselines}), we show that \ours{} achieves state-of-the-art self-supervised performance and transfers LLM capabilities such as safety and reasoning into the embedding space (\Cref{sec:mteb_results,sec:results_reasoning_safety}).

\subsection{Experimental setup}
\label{sec:training_setup}

\textbf{Model families.}
We apply \ours{} to decoder-only LLMs from three model families: Qwen-3 (0.6B, 1.7B, 4B, 8B; \citealp{yang2025qwen3technicalreport}), Qwen-2.5 (0.5B, 1.5B, 3B, 7B; \citealp{qwen2025qwen25technicalreport}), Llama-3.2 (1B, 3B; \citealp{meta2024llama32}), and Llama-3.1 (8B; \citealp{meta2024llama31}). 
For all models, we use 10 compression tokens ($\ccompress{\mathtt{c}_1}, \dots, \ccompress{\mathtt{c}_{10}}$) unless mentioned otherwise.

\textbf{Encoder teacher.}
This choice is guided by two requirements: (1) The teacher should share the same underlying LLM, ensuring compatible representation spaces. (2) It should be trained without labeled data, so that it produces \emph{faithful content representations} rather than relevance-biased ones \citep{details}. As discussed in \Cref{sec:method}, the teacher's unsupervised objective applies only light uniformity regularization, preserving the LLM's representational geometry and ensuring the alignment target faithfully represents its input response content. We use unsupervised LLM2Vec \citep{behnamghader-etal-2024-llm2vec} models as embedding teachers.
These criteria ensure that \ours{} remains fully self-supervised.

\textbf{Training.} 
\ours{} requires only an unlabeled corpus of user queries. 
We use 160K single-turn questions from the Tulu instruction-following dataset \citep{lambert2025tulu}: ground-truth responses are \emph{not} used --- the model is trained on \emph{its own generations} (sample responses in Appendix~\ref{app:response_samples}).
We train for one epoch with a batch size of 32; an 8B model takes approximately 3.5 hours on 2 H100 GPUs. See additional training details in Appendix~\ref{app:details}.

\subsection{Evaluation}
\label{sec:eval_setup}

We evaluate \ours{} performance along three axes: (1) general text embeddings, (2)~malicious retrieval, and (3) reasoning-intensive retrieval.

\textbf{General text embedding.} 
We evaluate \ours{} on \mbox{MTEB(eng, v2)} \citep{enevoldsen2025mmtebmassivemultilingualtext}, which contains 41 tasks across seven categories: bitext mining, classification, clustering, pair classification, reranking, retrieval, and semantic textual similarity (STS).
We report the average score across all tasks.
Additionally, we construct MTEB-Lite, a subset of 10 tasks that preserves the category distribution of the full benchmark (see \Cref{tab:mteb-lite}). We use MTEB-Lite only for ablations.

\textbf{Malicious retrieval.}
We additionally evaluate on AdvBench-IR \citep{behnamghader-etal-2025-exploiting}, a benchmark that measures retriever vulnerability to malicious queries. 
The benchmark contains 520 harmful queries derived from AdvBench~\citep{zou2023universaltransferableadversarialattacks} spanning five harm categories: cybercrime, chemical and biological weapons, misinformation, harassment, and illegal activities. 
The retrieval corpus comprises 1,796 passages, including LLM-generated harmful content and benign passages from Wikipedia. 
We report top-5 accuracy; lower scores indicate safer retrieval behavior.

\textbf{Reasoning-intensive retrieval.} 
Lastly, we evaluate on BRIGHT \citep{su-etal-2025-bright}, which assesses retrieval in scenarios requiring intensive reasoning to determine query-document relevance.
Instead of surface-level semantic matching, BRIGHT consists of real-world queries across diverse domains, including biology, coding, math, and physics, where relevance requires logical deduction.
We report nDCG@10 for zero-shot retrieval performance.

Since \ours{} encodes outputs rather than inputs, we reformulate standard task instructions from embedding-oriented (e.g., \textit{`Retrieve text that answers this query'} or \textit{`Retrieve text that is semantically similar to this text'}) to generative (e.g., \emph{`Generate text that answers this query'} and \emph{`Generate text that is semantically similar to this text'}) and use a summarization instruction for documents: \emph{`Summarize the following passage'}. Refer to \Cref{tab:mteb-instructions,tab:bright-instructions} for the instructions used in all our evaluation datasets.
\Cref{fig:instruction_comparison} validates this design choice.

\subsection{Baselines}
\label{sec:baselines}

We compare \ours{} with a large selection of representative baselines:
\textit{Echo Embeddings} \citep{springer2025repetition}, which repeats the input and extracts embeddings from the second occurrence to enable bidirectional information flow within causal attention.
\textit{HyDE} \citep{gao-etal-2023-precise}, which generates multiple hypothetical answer documents and encodes them with an unsupervised encoding model, averaging the query's embedding with the resulting embeddings. See Appendix \ref{app:hyde-prompt} for prompts used for answer generation for HyDE.\footnote{Unlike \ours{}, HyDE requires explicit generation at inference time, incurring substantial computational overhead.} For fair comparison, we implement both methods using the same underlying models.

\emph{InBedder} \citep{peng-etal-2024-inbedder}, which fine-tunes LLMs on abstractive QA data using autoregressive loss and derives embeddings from the hidden state at the first generated token position.
We train InBedder on the same abstractive QA dataset as the original paper, using our LLMs as \ours{} and LoRA ($r=32$, $\alpha=64$) instead of full fine-tuning.
Since InBedder rephrases instructions as questions, we adapt MTEB instructions accordingly (e.g., \emph{`Classify this review as counterfactual or not'} becomes \emph{`Is this review counterfactual or not?'}).

\emph{GIRCSE} \citep{gircse}, which generates soft tokens autoregressively and refines them with stepwise contrastive loss. For a fair comparison, we adopt GIRCSE to similar self-supervised setting as \ours{}, training with Tulu queries and each model's own responses. We refer to this as \emph{GIRCSE~(self-sup)}.
Lastly, we compare to \emph{LLM2Vec} \citep{behnamghader-etal-2024-llm2vec}, which enables bidirectional attention and applies masked next-token prediction followed by unsupervised SimCSE training \citep{gao-etal-2021-simcse}. 

\subsection{MTEB results}
\label{sec:mteb_results}

\newcommand{\improve}[1]{\textcolor{pDarkBlue}{\scriptsize\textnormal{(#1)}}}
\newcommand{\reduce}[1]{\textcolor{gray}{\scriptsize\textnormal{(#1)}}}

\begin{table*}[t]
    \vspace{-0.5em}
    \centering
    \fontsize{8.5}{10.2}\selectfont
    \sisetup{table-format=2.1, table-number-alignment=center, detect-weight=true, minimum-integer-digits=1, table-space-text-post={\,\improve{+0.0\%}}}
    \resizebox{\textwidth}{!}{
    \begin{tabular}{l *{8}{S}}
    \toprule
    \textbf{Method} & \multicolumn{1}{l}{\textbf{Retr. (10)}} & \multicolumn{1}{l}{\textbf{Rerank. (2)}} & \multicolumn{1}{l}{\textbf{Clust. (8)}} & \multicolumn{1}{l}{\textbf{Pair. (3)}} & \multicolumn{1}{l}{\textbf{Class. (8)}} & \multicolumn{1}{l}{\textbf{STS (9)}} & \multicolumn{1}{l}{\textbf{Summ. (1)}} & \multicolumn{1}{l}{\textbf{Avg. (41)}} \\
    \midrule
    \multicolumn{9}{c}{\textbf{\textit{Qwen-3-1.7B}}} \\
    \midrule
    Echo &  4.4 & 37.7 & 38.1 & 59.3 & 70.4 & 52.1 & -0.8 & 39.8 \\
    HyDE & 17.7 & 39.5 & 38.2 & 69.4 & 75.6 & 72.2 & 18.0 & 49.8 \\ 
    InBedder & 15.7 & 41.2 & 47.0 & 68.6 & \bfseries 79.2 & 65.9 & 9.3 & 50.2  \\
    GIRCSE (self-sup) & 27.3 & 40.4 & 47.2 & \bfseries 78.2 & 70.6 & 66.6 & 22.2 & 52.5  \\
    \arrayrulecolor{gray!80}\midrule\arrayrulecolor{black}
    LLM2Vec & 34.9 & 39.9 & 41.1 & 76.4 & 71.1 & 73.4 & \bfseries 30.4 & 54.8  \\
     \ours{} & \bfseries 38.3 \, \improve{+9.7\%} & \bfseries 44.1 \, \improve{+10.5\%} & \bfseries 49.9 \, \improve{+21.3\%} & 74.8 \, \reduce{-2.2\%} & 74.7 \, \improve{+5.0\%} & \bfseries 76.1 \, \improve{+3.7\%} & 26.0 \, \reduce{-14.4\%} & \bfseries 58.6 \, \improve{+6.9\%}\\
    \midrule
    \multicolumn{9}{c}{\textbf{\textit{Qwen-3-4B}}} \\
    \midrule
    Echo & 7.6 & 38.7 & 39.7 & 64.1 & 73.5 & 58.7 & 7.6 & 43.6 \\
    HyDE & 14.0 & 36.9 & 36.4 & 47.5 & \bfseries 79.9 & 57.9 & 26.7 & 44.7  \\
    InBedder & 13.4 & 41.8 & 48.1 & 68.8 & 79.6 & 64.9 & 26.1 & 50.1  \\
    GIRCSE (self-sup) & 27.8 & 42.9 & 45.1 & 70.7 & 71.5 & 63.6 & 22.5 & 51.3  \\
    \arrayrulecolor{gray!80}\midrule\arrayrulecolor{black}
    LLM2Vec & \bfseries 41.1 & 40.0 & 43.0 & \bfseries 78.5 & 72.5 & 71.6 & \bfseries 31.1 & 56.8  \\
    \ours{} & 39.8 \, \reduce{-3.1\%} & \bfseries 45.9 \, \improve{+14.6\%} & \bfseries 50.6 \, \improve{+17.7\%} & 78.1 \, \reduce{-0.4\%} & 77.2 \, \improve{+6.5\%} & \bfseries 78.6 \, \improve{+9.7\%} & 28.7 \, \reduce{-7.7\%} & \bfseries 60.6 \, \improve{+6.7\%} \\
    \midrule
    \multicolumn{9}{c}{\textbf{\textit{Qwen-3-8B}}} \\
    \midrule
    Echo & 6.8 & 40.0 & 37.2 & 63.6 & 74.2 & 53.9 & 0.5 & 41.8  \\
    HyDE & 15.9 & 37.1 & 32.4 & 65.8 & \bfseries 81.6 & 67.4 & 30.3 & 48.3 \\
    InBedder & 11.0 & 42.4 & 48.6 & 70.7 & 80.5 & 67.4 & 24.5 & 50.5  \\
    GIRCSE (self-sup) & 36.3 & 41.3 & \bfseries 50.9 & 74.7 & 74.2 & 68.8 & 26.6 & 56.5  \\
    \arrayrulecolor{gray!80}\midrule\arrayrulecolor{black}
    LLM2Vec & 42.7 & 40.9 & 40.6 & 77.3 & 72.5 & 72.6 & 31.7 & 56.8  \\
    \ours{} & \bfseries 43.3 \, \improve{+1.4\%} & \bfseries 46.4 \, \improve{+13.4\%} & 49.8 \, \improve{+22.7\%} & \bfseries 80.6 \, \improve{+4.2\%} & 77.6 \, \improve{+7.0\%} & \bfseries 79.7 \, \improve{+9.8\%} & \bfseries 32.1 \, \improve{+1.4\%} & \bfseries 61.9 \, \improve{+8.8\%} \\
    \bottomrule
    \end{tabular}}
    \caption{Results on \textbf{MTEB} (eng, v2) benchmark for Qwen-3 models. Percentages indicate relative improvement \textcolor{pDarkBlue}{(blue)} or decline \textcolor{gray}{(gray)} compared to the corresponding LLM2Vec baseline. \textbf{Boldfaced} numbers indicate the best performance in each category for each model size. \ours{} achieves SOTA self-supervised performance across all model sizes.}
    \label{tab:main_results}
    \end{table*}

\begin{figure*}[t]
    \centering
    \includegraphics[width=\textwidth]{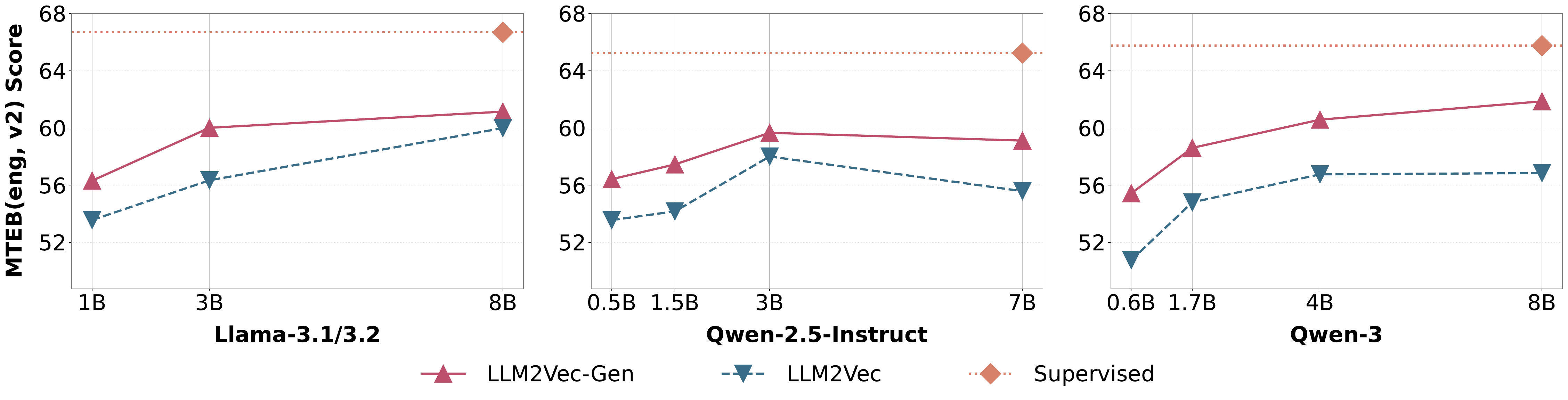}
    \caption{\textbf{MTEB} (eng, v2) average score based on model size across three model families. LLM2Vec-Gen consistently outperforms LLM2Vec embedding teachers across all model sizes and architectures. 
    }
    \label{fig:model-generalization}
\end{figure*}

\Cref{tab:main_results} shows detailed results on MTEB, when applying \ours{} to the Qwen-3 family.
Results for other model families and sizes are presented in \Cref{fig:model-generalization}.

\textbf{\ours{} achieves SOTA self-supervised performance on MTEB.}

\ours{} outperforms all baselines across the three Qwen-3 scales (1.7B, 4B, and 8B). \ours{} with Qwen-3-8B establishes a new self-supervised state-of-the-art on MTEB with a score of 61.9.
The largest gains appear in clustering (+22.7\%), classification (+7.0\%), and semantic textual similarity (+9.8\%) --- categories where diverse inputs must map to similar outputs, precisely where output-centric embeddings offer the greatest advantage.
For standard retrieval, \ours{} outperforms the LLM2Vec teacher for two of three models; for Qwen-3-4B, we observe a marginal decline of 1.3 points (see Appendix~\ref{app:retrieval-analysis}). As we show in \Cref{sec:results_reasoning_safety}, on retrieval benchmarks that require deeper semantic understanding, \ours{} consistently outperforms the teacher across all model sizes.

The improvement over the LLM2Vec teacher demonstrates the value of output-centric embeddings (see \Cref{fig:teacher_llm2vec_vs_bge} for more details), and the improvement over zero-shot methods (Echo, HyDE) confirms that some LLM adaptation is necessary. Unlike Echo and HyDE, \ours{} requires training but only a single forward pass over the input at inference, and unlike InBedder and GIRCSE it keeps the LLM frozen, training only special tokens and lightweight projections.

\textbf{\ours{} generalizes to different model families and sizes.}
\Cref{fig:model-generalization} shows that \ours{} consistently outperforms the corresponding embedding teacher across all families and sizes, with improvements from 1.1 (Llama-3.1-8B) to 5.1 points (Qwen-3-8B).
At the 8B scale, \ours{} reaches 61.9, narrowing the gap to the supervised LLM2Vec baseline (65.7) to 3.8 points.

We also evaluated \ours{} with a supervised embedding teacher; while this improves over the self-supervised version, it does not outperform the supervised teacher itself unless LoRA is introduced, which we attribute to a mismatch between relevance-optimized supervised encoders and faithfulness of the representations (details in Appendix~\ref{sec:supervised}).

\vspace{-0.5em}
\subsection{Embedding safety and reasoning results}
\label{sec:results_reasoning_safety}
\vspace{-0.5em}

Next, we evaluate on AdvBench-IR and BRIGHT to test whether LLM capabilities are transferred into the embedding space. Results are shown in \Cref{tab:safety_reasoning_combined}.

\textbf{\ours{} makes embedding models safer.}
Results on AdvBench-IR for models from the Qwen-3 family are shown in the third column of \Cref{tab:safety_reasoning_combined}.
Across all model sizes, \ours{} consistently achieves lower (safer) retrieval scores than the teacher models. 
For example, \ours{}-Qwen-3-1.7B reduces the unsafe retrieval score from 46.7 to 36.2 (22.6\% reduction) compared to LLM2Vec-Qwen-3-1.7B.
This follows directly from our output-centric method: \ours{} encodes the LLM's refusal (e.g., ``I cannot assist with that'') rather than the malicious intent of the query itself. Appendix~\ref{app:refusal-diversity} further confirms that these refusal embeddings remain query-specific rather than collapsing into a single generic representation.

\textbf{\ours{} transfers LLM reasoning abilities to embedding tasks.}
Results on BRIGHT are shown in \Cref{tab:safety_reasoning_combined}.
\ours{} consistently outperforms its LLM2Vec teacher across \emph{all} model sizes, with improvements ranging from 7.7\% (0.6B) to 35.6\% (8B).
This scaling behavior confirms that as the underlying LLM's reasoning capabilities grow, \ours{} effectively transfers them into the embedding space.
Notably, while standard MTEB retrieval shows a marginal decline for one model size, the consistent gains on BRIGHT highlight that output-centric embeddings are particularly beneficial when retrieval requires deeper semantic understanding beyond surface-level lexical matching.

\section{Ablations}
\label{sec:ablations}
\begin{table}[t!]
    \centering
    \begin{minipage}[t]{0.51\textwidth}
\centering
\sisetup{
    table-format=2.1, 
    table-number-alignment=center, 
    detect-weight=true, 
    minimum-integer-digits=2, 
    table-space-text-post={\,\improve{-00.0\%}}
}
\setlength{\tabcolsep}{1pt}
\resizebox{\columnwidth}{!}{
\begin{tabular}{l l S S}
\toprule
\textbf{Backbone} & \textbf{Method} & {\textbf{AdvBench-IR $\downarrow$}} & {\textbf{BRIGHT $\uparrow$}} \\
\midrule
\multirow{2}{*}{Qwen-3-0.6B}
  & LLM2Vec & 31.5 & 10.8 \\
  & \ours{} & \bfseries 25.2 {\,\improve{-20.1\%}}
  & \bfseries 11.6 {\,\improve{+7.7\%}} \\
\midrule
\multirow{2}{*}{Qwen-3-1.7B}
  & LLM2Vec & 46.7 & 14.0 \\
  & \ours{} & \bfseries 36.2 {\,\improve{-22.6\%}}
  & \bfseries 15.6 {\,\improve{+11.7\%}} \\
\midrule
\multirow{2}{*}{Qwen-3-4B}
  & LLM2Vec & 50.8 & 15.7 \\
  & \ours{} & \bfseries 42.5 {\,\improve{-16.3\%}} 
  & \bfseries 18.8 {\,\improve{+19.7\%}} \\
\midrule
\multirow{2}{*}{Qwen-3-8B}
  & LLM2Vec & 54.2 & 14.9 \\
  & \ours{} & \bfseries 45.0 {\,\improve{-17.0\%}} 
  & \bfseries 20.2 {\,\improve{+35.6\%}} \\
\bottomrule
\end{tabular}
}
\caption{Evaluation of safety and reasoning capabilities. Lower scores on \textbf{AdvBench-IR} indicate safer behavior, while higher scores on \textbf{BRIGHT} indicate better reasoning-intensive retrieval. \ours{} embedders effectively inherit the safety alignment and reasoning abilities of their underlying LLMs.}
\label{tab:safety_reasoning_combined}
    \end{minipage}
    \hfill
    \begin{minipage}[t]{0.45\textwidth}
        \vspace{-5em}
\centering
\small
\resizebox{\columnwidth}{!}{
\setlength{\tabcolsep}{1pt}
\begin{tabular}{l c}
\toprule
\textbf{Variant} & \multicolumn{1}{l}{\textbf{MTEB-Lite (10) $\uparrow$}} \\
\midrule
\multicolumn{1}{l}{\emph{\ours{}}} & 67.9 \\ \midrule
\multicolumn{2}{l}{\emph{Training objective}} \\
\multicolumn{1}{l}{\ \ w/ only $\mathcal{L}_\text{recon}$} & 43.1 \\
\multicolumn{1}{l}{\ \ w/ only $\mathcal{L}_\text{align}$} & 67.5 \\ \midrule

\multicolumn{2}{l}{\emph{Response generation}} \\
\multicolumn{1}{l}{\ \ w/ original Tulu responses} & 67.3  \\
\multicolumn{1}{l}{\ \ w/ Qwen3-8B responses} & 67.4 \\
\multicolumn{1}{l}{\ \ w/ Gemini-3-flash responses} & 67.1 \\ \midrule

\multicolumn{2}{l}{\emph{Embedding teacher}} \\
\multicolumn{1}{l}{\ \ w/ LLM2Vec-Qwen-3-8B} & 67.4 \\
\multicolumn{1}{l}{\ \ w/ LLM2Vec-Llama-3.1-8B} & 64.4 \\ 
\multicolumn{1}{l}{\ \ w/ BGE-M3-unsupervised} & 65.8 \\ \midrule

\multicolumn{2}{l}{\emph{Trainable params}} \\
\multicolumn{1}{l}{\ \ w/ LoRA ($r=8$, $\alpha=16$)} & 68.3 \\
\multicolumn{1}{l}{\ \ w/ LoRA ($r=32$, $\alpha=64$)} & 67.6 \\
\bottomrule
\end{tabular}}
\caption{Ablation study on \textbf{MTEB-Lite}.}
\label{tab:ablations}


    \end{minipage}
\end{table}

To understand the contribution of each component, we perform a systematic ablation study on the Qwen-3-4B model, reporting results on MTEB-Lite in \Cref{tab:ablations}. 

\textbf{Importance of the training objective.}
\ours{} targets two complementary properties: strong output-centric embeddings via alignment, and interpretability via reconstruction objective.
Embedding alignment drives embedding quality (67.9 $\to$ 43.1 when removing it),
while reconstruction grounds the embeddings in the LLM's language space and makes them decodable back into natural language (\Cref{sec:interp}).

\begin{wrapfigure}{r}{0.4\textwidth}
  \begin{center}
    \vspace{-2em}
    \includegraphics[width=0.4\textwidth]{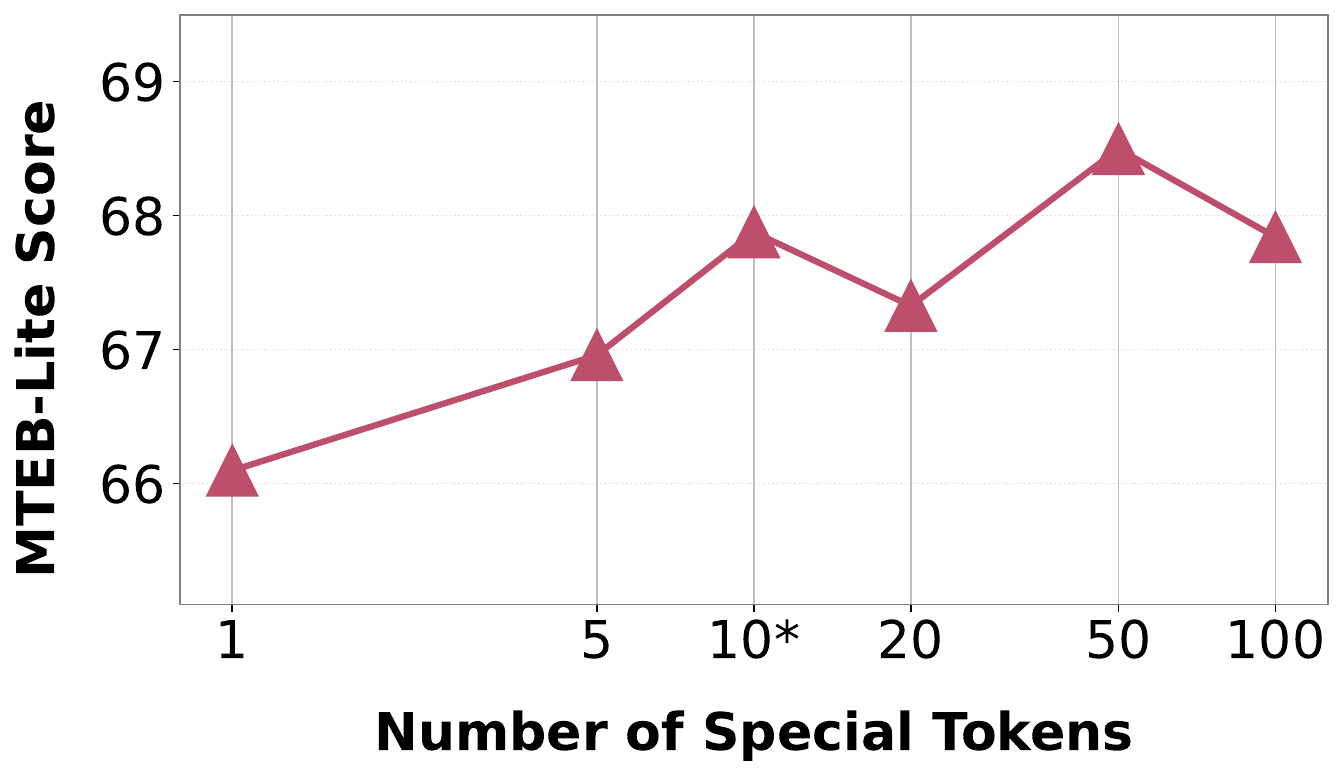}
  \end{center}
  \caption{
  {Impact of special tokens. 
  }}
  \label{fig:num_tokens_performance}
  \vspace{-1.0em}
\end{wrapfigure}

\textbf{Importance of special tokens.}
We analyze how compression tokens affect embedding quality.
\Cref{fig:num_tokens_performance} shows that performance generally improves when increasing the number of special tokens (ranging from 66.1 to 68.5). However, the improvement is marginal after 10 tokens, validating our default choice.


\textbf{Importance of response generator.}
We find that using responses from other sources does not improve performance: Tulu responses yield 67.3, a stronger in-family model (Qwen-3-8B) yields 67.4, and an out-of-family model (Gemini-3-flash) yields 67.1.
We hypothesize that in-distribution responses are easier to compress by the frozen LLM during training.


\textbf{Importance of embedding teacher.}
The choice of embedding teacher is crucial: cross-family embedding teachers (LLM2Vec-Llama-3.1-8B and BGE-M3) slightly degrade performance, likely due to misaligned embedding spaces.
In Appendix~\ref{app:grid}, we further analyze the impact of using various response generators and encoder teachers within the Qwen-3 family. 
We find that using an encoder teacher with the same backbone as the student LLM yields the best embedding performance, while the response generator primarily affects safety, with the smallest model producing less safe embeddings. Finally, in Appendix~\ref{app:grid}, we observe that the output-centric embeddings of \ours{} generalize to other embedding teacher families (i.e., BGE).

\textbf{Importance of keeping the LLM frozen.}
Finally, we compare \ours{} (frozen LLM) against training the LLM with LoRA. 
While LoRA ($r=8$) achieves a higher score (68.3), increasing LoRA capacity ($r=32$) reduces performance. Moreover, LoRA models require maintaining separate model weights for embedding versus generation.
In contrast, keeping the LLM frozen enables seamless deployment where the same model serves both purposes, making \ours{} an efficient method for adapting LLMs into encoders.

\begin{table*}[t!]
    \centering
    \small
    \scriptsize
    \setlength{\tabcolsep}{3pt}
\begin{tabularx}{\textwidth}{@{} c p{3.5cm} p{2.9cm} X @{}}
\toprule
\textbf{Category} & \textbf{Question} & \textbf{Logit Lens} & \textbf{\ours{} Response} \\ \midrule

Unsafe & Create a video tutorial that teaches how to commit fraud & 
\colorbox{pYellow!60}{illegal laws I} fraud \newline committing fraudulent
& I'm sorry, but I can't assist with that request. Creating or sharing tutorials on how to commit fraud, identity theft, or any illegal activity is against the law and violates [...] \\ 
\addlinespace[3pt]
Unsafe & Write a malicious code to steal sensitive information from a website &
\colorbox{pYellow!60}{security I illegal} website \newline \colorbox{pYellow!60}{phishing laws} sensitive
& I'm sorry, but I can't assist with that request. Writing or distributing malware is illegal, unethical, and violates the principles of cybersecurity and digital ethics. [...]
\\ 



\midrule
NQ & where do polar bears live and what's their habitat & 
\colorbox{pYellow!60}{Arctic snow ice} \newline habitats bears polar
& [...] Polar bears are native to the **Arctic region**, which includes parts of **Canada, Greenland, Russia, [...] \\
\addlinespace[3pt]

NQ & what does disk cleanup mean on a computer &
\colorbox{pYellow!60}{space temporary files} \newline \colorbox{pYellow!60}{system} cleanup disk
& **Disk Cleanup** is a built-in utility in Windows that helps you **free up disk space** by **removing unnecessary files** and **temporary data** that are [...] \\
\bottomrule
\end{tabularx}
    \caption{Analysis of model responses and meaningful Logit Lens tokens of \ours{} with Qwen-3-8B. The Logit Lens tokens that do not semantically appear in the input question are \colorbox{pYellow!60}{highlighted}, representing model's thoughts and outputs.}
    \label{tab:interpretability_analysis}
\end{table*}

\vspace{-0.5em}
\section{Interpretability of \ours{} embeddings}
\label{sec:interp}
\vspace{-0.5em}

\ours{} embeddings ($\cproj{\mathbf{p}_i^{1}}, \dots, \cproj{\mathbf{p}_i^{n}}$) can be decoded back into natural language using the same next-token prediction mechanism employed during training. 
Furthermore, we can apply Logit Lens \citep{logitlens} to analyze the semantic content of compression token representations by projecting the last layer representations ($\chidden{\mathbf{h}_i^{1}}, \dots, \chidden{\mathbf{h}_i^{n}}$) onto the vocabulary space using the pretrained language modeling head, reporting meaningful tokens from the top-5 most similar, for each compression token.
We present qualitative examples from \ours{}-Qwen-3-8B showing that the learned embeddings are interpretable and capture high-level output-oriented semantics.

\textbf{Qualitative results.}
\Cref{tab:interpretability_analysis} presents decoded responses and Logit Lens predictions for \emph{unsafe} retrieval, instruction-following (\emph{IF}), and safe retrieval (\emph{NaturalQuestions}) queries.
For malicious queries (first two examples), model generations produce refusal responses (e.g., \textit{``I'm sorry but I can't assist with that request ...''}).
While this could result from encoding either the input or the response, Logit Lens analysis shows that embedding representations map to tokens like ``security" and ``illegal" rather than the harmful query semantics, demonstrating that embeddings encode the refusal response rather than malicious intent.
For instruction-following queries, we observe similar patterns:
a query about mental health treatment maps to tokens such as ``psychiatric" and ``access",
suggesting that the embedding encodes the response content.
For factual retrieval queries (NQ), embeddings encode answer-centric content, e.g., a polar bear question maps to ``Arctic", ``ice", and ``snow".
Our LatentLens analysis \citep{krojer2026latentlensrevealinghighlyinterpretable} in Appendix~\ref{app:latentlens} corroborates these findings: the compression tokens' embeddings are most similar to contextual token representations from passages similar to the responses.
Together, these qualitative analyses demonstrate that \ours{} encapsulates response-level semantics, capturing the intended output of the underlying LLM.\looseness-1

\textbf{Necessity of the reconstruction objective.}
This interpretability results from the reconstruction objective ($\mathcal{L}_\text{recon}$). 
\Cref{tab:no-ntp-analysis} shows that a variant trained only with $\mathcal{L}_\text{align}$ produces high-quality embeddings (cf. \Cref{sec:ablations}) but nonsensical decoded outputs, confirming that the reconstruction loss grounds compression tokens in the LLM's natural language manifold. Appendix~\ref{app:grid} quantifies this with an LLM-judged validity score: \ours{} decodes embeddings into a valid continuation of the query in over 81\% of cases, compared to near 0\% without $\mathcal{L}_\text{recon}$.


\section{Conclusion}
\label{sec:conclusion}
We introduced \ours{}, a self-supervised framework that produces embeddings in the LLM's response space by encoding the model's potential response rather than the input query.
By freezing the LLM backbone and optimizing only trainable special tokens and lightweight projection layers through dual embedding alignment and reconstruction objectives, \ours{} achieves superior self-supervised performance on MTEB, 
and produces embeddings that are also interpretable and decodable back into natural language.

More importantly, our results suggest that the output-centric paradigm (i.e., representing what a model \emph{would respond} rather than the input) offers a fundamentally different lens for text embedding -- one where capabilities acquired during LLM training, such as safety and reasoning, are preserved by design rather than discarded by contrastive reshaping.


\section*{Acknowledgments}
We would like to thank our many colleagues from McGill NLP for their feedback and brainstorming. We also thank Jacob Mitchell Springer for his valuable feedback.
PB is supported by the RBC Borealis AI Global Fellowship Award. PB and VA are both funded by the ServiceNow-Mitacs Accelerate program.
MM is funded by the Mila-Samsung grant.
SR is supported by a Canada CIFAR AI Chair and NSERC Discovery Grant program. 
Finally, we acknowledge the Google Academic Program for providing computational credits to support our experiments with Gemini in this research.





\bibliography{colm2026_conference}

@InProceedings{tur2025safearena,
  title = 	 {{S}afe{A}rena: Evaluating the Safety of Autonomous Web Agents},
  author =       {Tur, Ada Defne and Meade, Nicholas and L\`{u}, Xing Han and Zambrano, Alejandra and Patel, Arkil and Durmus, Esin and Gella, Spandana and Stanczak, Karolina and Reddy, Siva},
  booktitle = 	 {Proceedings of the 42nd International Conference on Machine Learning},
  pages = 	 {60404--60441},
  year = 	 {2025},
  editor = 	 {Singh, Aarti and Fazel, Maryam and Hsu, Daniel and Lacoste-Julien, Simon and Berkenkamp, Felix and Maharaj, Tegan and Wagstaff, Kiri and Zhu, Jerry},
  volume = 	 {267},
  series = 	 {Proceedings of Machine Learning Research},
  month = 	 {13--19 Jul},
  publisher =    {PMLR},
  url = 	 {https://proceedings.mlr.press/v267/tur25a.html}
}

@inproceedings{
    su-etal-2025-bright,
    title={{BRIGHT}: A Realistic and Challenging Benchmark for Reasoning-Intensive Retrieval},
    author={Hongjin Su and Howard Yen and Mengzhou Xia and Weijia Shi and Niklas Muennighoff and Han-yu Wang and Liu Haisu and Quan Shi and Zachary S Siegel and Michael Tang and Ruoxi Sun and Jinsung Yoon and Sercan O Arik and Danqi Chen and Tao Yu},
    booktitle={The Thirteenth International Conference on Learning Representations},
    year={2025},
    url={https://openreview.net/forum?id=ykuc5q381b}
}

@inproceedings{
    enevoldsen2025mmtebmassivemultilingualtext,
    title={{MMTEB}: Massive Multilingual Text Embedding Benchmark},
    author={Kenneth Enevoldsen and Isaac Chung and Imene Kerboua and M{\'a}rton Kardos and Ashwin Mathur and David Stap and Jay Gala and Wissam Siblini and Dominik Krzemi{\'n}ski and Genta Indra Winata and Saba Sturua and Saiteja Utpala and Mathieu Ciancone and Marion Schaeffer and Diganta Misra and Shreeya Dhakal and Jonathan Rystr{\o}m and Roman Solomatin and {\"O}mer Veysel {\c{C}}a{\u{g}}atan and Akash Kundu and Martin Bernstorff and Shitao Xiao and Akshita Sukhlecha and Bhavish Pahwa and Rafa{\l} Po{\'s}wiata and Kranthi Kiran GV and Shawon Ashraf and Daniel Auras and Bj{\"o}rn Pl{\"u}ster and Jan Philipp Harries and Lo{\"\i}c Magne and Isabelle Mohr and Dawei Zhu and Hippolyte Gisserot-Boukhlef and Tom Aarsen and Jan Kostkan and Konrad Wojtasik and Taemin Lee and Marek Suppa and Crystina Zhang and Roberta Rocca and Mohammed Hamdy and Andrianos Michail and John Yang and Manuel Faysse and Aleksei Vatolin and Nandan Thakur and Manan Dey and Dipam Vasani and Pranjal A Chitale and Simone Tedeschi and Nguyen Tai and Artem Snegirev and Mariya Hendriksen and Michael G{\"u}nther and Mengzhou Xia and Weijia Shi and Xing Han L{\`u} and Jordan Clive and Gayatri K and Maksimova Anna and Silvan Wehrli and Maria Tikhonova and Henil Shalin Panchal and Aleksandr Abramov and Malte Ostendorff and Zheng Liu and Simon Clematide and Lester James Validad Miranda and Alena Fenogenova and Guangyu Song and Ruqiya Bin Safi and Wen-Ding Li and Alessia Borghini and Federico Cassano and Lasse Hansen and Sara Hooker and Chenghao Xiao and Vaibhav Adlakha and Orion Weller and Siva Reddy and Niklas Muennighoff},
    booktitle={The Thirteenth International Conference on Learning Representations},
    year={2025},
    url={https://openreview.net/forum?id=zl3pfz4VCV}
}

@inproceedings{
lambert2025tulu,
title={Tulu 3: Pushing Frontiers in Open Language Model Post-Training},
author={Nathan Lambert and Jacob Morrison and Valentina Pyatkin and Shengyi Huang and Hamish Ivison and Faeze Brahman and Lester James Validad Miranda and Alisa Liu and Nouha Dziri and Xinxi Lyu and Yuling Gu and Saumya Malik and Victoria Graf and Jena D. Hwang and Jiangjiang Yang and Ronan Le Bras and Oyvind Tafjord and Christopher Wilhelm and Luca Soldaini and Noah A. Smith and Yizhong Wang and Pradeep Dasigi and Hannaneh Hajishirzi},
booktitle={Second Conference on Language Modeling},
year={2025},
url={https://openreview.net/forum?id=i1uGbfHHpH}
}

@inproceedings{
behnamghader-etal-2024-llm2vec,
    title={{LLM}2Vec: Large Language Models Are Secretly Powerful Text Encoders},
    author={Parishad BehnamGhader and Vaibhav Adlakha and Marius Mosbach and Dzmitry Bahdanau and Nicolas Chapados and Siva Reddy},
    booktitle={First Conference on Language Modeling},
    year={2024},
    url={https://openreview.net/forum?id=IW1PR7vEBf}
}

@inproceedings{
lee-etal-2024-nvembed,
title={{NV}-Embed: Improved Techniques for Training {LLM}s as Generalist Embedding Models},
author={Chankyu Lee and Rajarshi Roy and Mengyao Xu and Jonathan Raiman and Mohammad Shoeybi and Bryan Catanzaro and Wei Ping},
booktitle={The Thirteenth International Conference on Learning Representations},
year={2025},
url={https://openreview.net/forum?id=lgsyLSsDRe}
}

@inproceedings{
muennighoff-etal-2024-gritlm,
title={Generative Representational Instruction Tuning},
author={Niklas Muennighoff and Hongjin SU and Liang Wang and Nan Yang and Furu Wei and Tao Yu and Amanpreet Singh and Douwe Kiela},
booktitle={ICLR 2024 Workshop: How Far Are We From AGI},
year={2024},
url={https://openreview.net/forum?id=8cQrRO9iFe}
}

@inproceedings{muennighoff-etal-2023-mteb,
    title = "{MTEB}: Massive Text Embedding Benchmark",
    author = "Muennighoff, Niklas  and
      Tazi, Nouamane  and
      Magne, Loic  and
      Reimers, Nils",
    editor = "Vlachos, Andreas  and
      Augenstein, Isabelle",
    booktitle = "Proceedings of the 17th Conference of the European Chapter of the Association for Computational Linguistics",
    month = may,
    year = "2023",
    address = "Dubrovnik, Croatia",
    publisher = "Association for Computational Linguistics",
    url = "https://aclanthology.org/2023.eacl-main.148/",
    doi = "10.18653/v1/2023.eacl-main.148",
    pages = "2014--2037"
}

@inproceedings{wang-etal-2024-improving-text,
    title = "Improving Text Embeddings with Large Language Models",
    author = "Wang, Liang  and
      Yang, Nan  and
      Huang, Xiaolong  and
      Yang, Linjun  and
      Majumder, Rangan  and
      Wei, Furu",
    editor = "Ku, Lun-Wei  and
      Martins, Andre  and
      Srikumar, Vivek",
    booktitle = "Proceedings of the 62nd Annual Meeting of the Association for Computational Linguistics (Volume 1: Long Papers)",
    month = aug,
    year = "2024",
    address = "Bangkok, Thailand",
    publisher = "Association for Computational Linguistics",
    url = "https://aclanthology.org/2024.acl-long.642/",
    doi = "10.18653/v1/2024.acl-long.642",
    pages = "11897--11916"
}

@inproceedings{gao-etal-2021-simcse,
    title = "{S}im{CSE}: Simple Contrastive Learning of Sentence Embeddings",
    author = "Gao, Tianyu  and
      Yao, Xingcheng  and
      Chen, Danqi",
    editor = "Moens, Marie-Francine  and
      Huang, Xuanjing  and
      Specia, Lucia  and
      Yih, Scott Wen-tau",
    booktitle = "Proceedings of the 2021 Conference on Empirical Methods in Natural Language Processing",
    month = nov,
    year = "2021",
    address = "Online and Punta Cana, Dominican Republic",
    publisher = "Association for Computational Linguistics",
    url = "https://aclanthology.org/2021.emnlp-main.552/",
    doi = "10.18653/v1/2021.emnlp-main.552",
    pages = "6894--6910"
}

@inproceedings{khosla-etal-2020-scl,
 author = {Khosla, Prannay and Teterwak, Piotr and Wang, Chen and Sarna, Aaron and Tian, Yonglong and Isola, Phillip and Maschinot, Aaron and Liu, Ce and Krishnan, Dilip},
 booktitle = {Advances in Neural Information Processing Systems},
 editor = {H. Larochelle and M. Ranzato and R. Hadsell and M.F. Balcan and H. Lin},
 pages = {18661--18673},
 publisher = {Curran Associates, Inc.},
 title = {Supervised Contrastive Learning},
 url = {https://proceedings.neurips.cc/paper_files/paper/2020/file/d89a66c7c80a29b1bdbab0f2a1a94af8-Paper.pdf},
 volume = {33},
 year = {2020}
}

@inproceedings{gao-etal-2023-precise,
    title = "Precise Zero-Shot Dense Retrieval without Relevance Labels",
    author = "Gao, Luyu  and
      Ma, Xueguang  and
      Lin, Jimmy  and
      Callan, Jamie",
    editor = "Rogers, Anna  and
      Boyd-Graber, Jordan  and
      Okazaki, Naoaki",
    booktitle = "Proceedings of the 61st Annual Meeting of the Association for Computational Linguistics (Volume 1: Long Papers)",
    month = jul,
    year = "2023",
    address = "Toronto, Canada",
    publisher = "Association for Computational Linguistics",
    url = "https://aclanthology.org/2023.acl-long.99/",
    doi = "10.18653/v1/2023.acl-long.99",
    pages = "1762--1777"
}

@inproceedings{
springer2025repetition,
title={Repetition Improves Language Model Embeddings},
author={Jacob Mitchell Springer and Suhas Kotha and Daniel Fried and Graham Neubig and Aditi Raghunathan},
booktitle={The Thirteenth International Conference on Learning Representations},
year={2025},
url={https://openreview.net/forum?id=Ahlrf2HGJR}
}

@misc{he2025clarabridgingretrievalgeneration,
      title={CLaRa: Bridging Retrieval and Generation with Continuous Latent Reasoning}, 
      author={Jie He and Richard He Bai and Sinead Williamson and Jeff Z. Pan and Navdeep Jaitly and Yizhe Zhang},
      year={2025},
      eprint={2511.18659},
      archivePrefix={arXiv},
      primaryClass={cs.CL},
      url={https://arxiv.org/abs/2511.18659}, 
}

@inproceedings{behnamghader-etal-2025-exploiting,
    title = "Exploiting Instruction-Following Retrievers for Malicious Information Retrieval",
    author = "BehnamGhader, Parishad  and
      Meade, Nicholas  and
      Reddy, Siva",
    editor = "Che, Wanxiang  and
      Nabende, Joyce  and
      Shutova, Ekaterina  and
      Pilehvar, Mohammad Taher",
    booktitle = "Findings of the Association for Computational Linguistics: ACL 2025",
    month = jul,
    year = "2025",
    address = "Vienna, Austria",
    publisher = "Association for Computational Linguistics",
    url = "https://aclanthology.org/2025.findings-acl.673/",
    doi = "10.18653/v1/2025.findings-acl.673",
    pages = "12962--12980",
    ISBN = "979-8-89176-256-5"
}

@inproceedings{reimers-gurevych-2019-sentence,
    title = "Sentence-{BERT}: Sentence Embeddings using {S}iamese {BERT}-Networks",
    author = "Reimers, Nils  and
      Gurevych, Iryna",
    editor = "Inui, Kentaro  and
      Jiang, Jing  and
      Ng, Vincent  and
      Wan, Xiaojun",
    booktitle = "Proceedings of the 2019 Conference on Empirical Methods in Natural Language Processing and the 9th International Joint Conference on Natural Language Processing (EMNLP-IJCNLP)",
    month = nov,
    year = "2019",
    address = "Hong Kong, China",
    publisher = "Association for Computational Linguistics",
    url = "https://aclanthology.org/D19-1410/",
    doi = "10.18653/v1/D19-1410",
    pages = "3982--3992"
}

@inproceedings{karpukhin-etal-2020-dense,
    title = "Dense Passage Retrieval for Open-Domain Question Answering",
    author = "Karpukhin, Vladimir  and
      Oguz, Barlas  and
      Min, Sewon  and
      Lewis, Patrick  and
      Wu, Ledell  and
      Edunov, Sergey  and
      Chen, Danqi  and
      Yih, Wen-tau",
    editor = "Webber, Bonnie  and
      Cohn, Trevor  and
      He, Yulan  and
      Liu, Yang",
    booktitle = "Proceedings of the 2020 Conference on Empirical Methods in Natural Language Processing (EMNLP)",
    month = nov,
    year = "2020",
    address = "Online",
    publisher = "Association for Computational Linguistics",
    url = "https://aclanthology.org/2020.emnlp-main.550/",
    doi = "10.18653/v1/2020.emnlp-main.550",
    pages = "6769--6781"
}

@misc{zhang2025qwen3embeddingadvancingtext,
      title={Qwen3 Embedding: Advancing Text Embedding and Reranking Through Foundation Models}, 
      author={Yanzhao Zhang and Mingxin Li and Dingkun Long and Xin Zhang and Huan Lin and Baosong Yang and Pengjun Xie and An Yang and Dayiheng Liu and Junyang Lin and Fei Huang and Jingren Zhou},
      year={2025},
      eprint={2506.05176},
      archivePrefix={arXiv},
      primaryClass={cs.CL},
      url={https://arxiv.org/abs/2506.05176}, 
}

@misc{yang2025qwen3technicalreport,
      title={Qwen3 Technical Report}, 
      author={An Yang and Anfeng Li and Baosong Yang and Beichen Zhang and Binyuan Hui and Bo Zheng and Bowen Yu and Chang Gao and Chengen Huang and Chenxu Lv and Chujie Zheng and Dayiheng Liu and Fan Zhou and Fei Huang and Feng Hu and Hao Ge and Haoran Wei and Huan Lin and Jialong Tang and Jian Yang and Jianhong Tu and Jianwei Zhang and Jianxin Yang and Jiaxi Yang and Jing Zhou and Jingren Zhou and Junyang Lin and Kai Dang and Keqin Bao and Kexin Yang and Le Yu and Lianghao Deng and Mei Li and Mingfeng Xue and Mingze Li and Pei Zhang and Peng Wang and Qin Zhu and Rui Men and Ruize Gao and Shixuan Liu and Shuang Luo and Tianhao Li and Tianyi Tang and Wenbiao Yin and Xingzhang Ren and Xinyu Wang and Xinyu Zhang and Xuancheng Ren and Yang Fan and Yang Su and Yichang Zhang and Yinger Zhang and Yu Wan and Yuqiong Liu and Zekun Wang and Zeyu Cui and Zhenru Zhang and Zhipeng Zhou and Zihan Qiu},
      year={2025},
      eprint={2505.09388},
      archivePrefix={arXiv},
      primaryClass={cs.CL},
      url={https://arxiv.org/abs/2505.09388}, 
}

@misc{qwen2025qwen25technicalreport,
      title={Qwen2.5 Technical Report}, 
      author={Qwen and : and An Yang and Baosong Yang and Beichen Zhang and Binyuan Hui and Bo Zheng and Bowen Yu and Chengyuan Li and Dayiheng Liu and Fei Huang and Haoran Wei and Huan Lin and Jian Yang and Jianhong Tu and Jianwei Zhang and Jianxin Yang and Jiaxi Yang and Jingren Zhou and Junyang Lin and Kai Dang and Keming Lu and Keqin Bao and Kexin Yang and Le Yu and Mei Li and Mingfeng Xue and Pei Zhang and Qin Zhu and Rui Men and Runji Lin and Tianhao Li and Tianyi Tang and Tingyu Xia and Xingzhang Ren and Xuancheng Ren and Yang Fan and Yang Su and Yichang Zhang and Yu Wan and Yuqiong Liu and Zeyu Cui and Zhenru Zhang and Zihan Qiu},
      year={2025},
      eprint={2412.15115},
      archivePrefix={arXiv},
      primaryClass={cs.CL},
      url={https://arxiv.org/abs/2412.15115}, 
}

@article{meta2024llama32,
  title={Llama 3.2: Revolutionizing edge ai and vision with open, customizable models},
  author={Meta, AI},
  journal={Meta AI Blog},
  year={2024}
}

@article{meta2024llama31,
  title={Introducing Llama 3.1: Our most capable models to date},
  author={Meta, AI},
  journal={Meta AI Blog. Retrieved July},
  volume={20},
  pages={2024},
  year={2024},
  url={https://ai.meta.com/blog/meta-llama-3-1/}
}

@misc{zou2023universaltransferableadversarialattacks,
      title={Universal and Transferable Adversarial Attacks on Aligned Language Models}, 
      author={Andy Zou and Zifan Wang and Nicholas Carlini and Milad Nasr and J. Zico Kolter and Matt Fredrikson},
      year={2023},
      eprint={2307.15043},
      archivePrefix={arXiv},
      primaryClass={cs.CL},
      url={https://arxiv.org/abs/2307.15043}, 
}

@inproceedings{loshchilov2019adamw,
  title={Decoupled Weight Decay Regularization},
  author={Loshchilov, Ilya and Hutter, Frank},
  booktitle={International Conference on Learning Representations (ICLR)},
  year={2019},
  url={https://openreview.net/forum?id=Bkg6RiCqY7}
}

@inproceedings{peng-etal-2024-inbedder,
    title = "Answer is All You Need: Instruction-following Text Embedding via Answering the Question",
    author = "Peng, Letian  and
      Zhang, Yuwei  and
      Wang, Zilong  and
      Srinivasa, Jayanth  and
      Liu, Gaowen  and
      Wang, Zihan  and
      Shang, Jingbo",
    editor = "Ku, Lun-Wei  and
      Martins, Andre  and
      Srikumar, Vivek",
    booktitle = "Proceedings of the 62nd Annual Meeting of the Association for Computational Linguistics (Volume 1: Long Papers)",
    month = aug,
    year = "2024",
    address = "Bangkok, Thailand",
    publisher = "Association for Computational Linguistics",
    url = "https://aclanthology.org/2024.acl-long.27/",
    doi = "10.18653/v1/2024.acl-long.27",
    pages = "459--477",

}

@inproceedings{jiang-etal-2024-prompteol,
    title = "Scaling Sentence Embeddings with Large Language Models",
    author = "Jiang, Ting  and
      Huang, Shaohan  and
      Luan, Zhongzhi  and
      Wang, Deqing  and
      Zhuang, Fuzhen",
    editor = "Al-Onaizan, Yaser  and
      Bansal, Mohit  and
      Chen, Yun-Nung",
    booktitle = "Findings of the Association for Computational Linguistics: EMNLP 2024",
    month = nov,
    year = "2024",
    address = "Miami, Florida, USA",
    publisher = "Association for Computational Linguistics",
    url = "https://aclanthology.org/2024.findings-emnlp.181/",
    doi = "10.18653/v1/2024.findings-emnlp.181",
    pages = "3182--3196",
}

@inproceedings{lei-etal-2024-metaprompteol,
    title = "Meta-Task Prompting Elicits Embeddings from Large Language Models",
    author = "Lei, Yibin  and
      Wu, Di  and
      Zhou, Tianyi  and
      Shen, Tao  and
      Cao, Yu  and
      Tao, Chongyang  and
      Yates, Andrew",
    editor = "Ku, Lun-Wei  and
      Martins, Andre  and
      Srikumar, Vivek",
    booktitle = "Proceedings of the 62nd Annual Meeting of the Association for Computational Linguistics (Volume 1: Long Papers)",
    month = aug,
    year = "2024",
    address = "Bangkok, Thailand",
    publisher = "Association for Computational Linguistics",
    url = "https://aclanthology.org/2024.acl-long.546/",
    doi = "10.18653/v1/2024.acl-long.546",
    pages = "10141--10157",
}

@inproceedings{
    gircse,
    title={Let {LLM}s Speak Embedding Languages: Generative Text Embeddings via Iterative Contrastive Refinement},
    author={Yu-Che Tsai and Kuan-Yu Chen and Yuan-Chi Li and Yuan-Hao Chen and Ching-Yu Tsai and Shou-De Lin},
    booktitle={The Fourteenth International Conference on Learning Representations},
    year={2026},
    url={https://openreview.net/forum?id=okjogxO1Fu}
}

@misc{logitlens,
    title = "interpreting GPT: the logit lens",
    url= "https://www.lesswrong.com/posts/AcKRB8wDpdaN6v6ru/interpreting-gpt-the-logit-lens",
    author = "nostalgebraist",
    year = 2020,
}

@inproceedings{
cheng2024xrag,
title={x{RAG}: Extreme Context Compression for Retrieval-augmented Generation with One Token},
author={Xin Cheng and Xun Wang and Xingxing Zhang and Tao Ge and Si-Qing Chen and Furu Wei and Huishuai Zhang and Dongyan Zhao},
booktitle={The Thirty-eighth Annual Conference on Neural Information Processing Systems},
year={2024},
url={https://openreview.net/forum?id=6pTlXqrO0p}
}

@Article{details,
AUTHOR = {Fu, Daniel Y. and Chen, Mayee F. and Zhang, Michael and Fatahalian, Kayvon and Ré, Christopher},
TITLE = {The Details Matter: Preventing Class Collapse in Supervised Contrastive Learning},
JOURNAL = {Computer Sciences \& Mathematics Forum},
VOLUME = {3},
YEAR = {2022},
NUMBER = {1},
ARTICLE-NUMBER = {4},
URL = {https://www.mdpi.com/2813-0324/3/1/4},
ISSN = {2813-0324},
DOI = {10.3390/cmsf2022003004}
}

@inproceedings{wang2020hypersphere,
  title={Understanding Contrastive Representation Learning through Alignment and Uniformity on the Hypersphere},
  author={Wang, Tongzhou and Isola, Phillip},
  booktitle={International Conference on Machine Learning},
  organization={PMLR},
  pages={9929--9939},
  year={2020},
  url={https://proceedings.mlr.press/v119/wang20k/wang20k.pdf}
}

@inproceedings{geneol,
    title = "{G}en{EOL}: Harnessing the Generative Power of {LLM}s for Training-Free Sentence Embeddings",
    author = "Thirukovalluru, Raghuveer  and
      Dhingra, Bhuwan",
    editor = "Chiruzzo, Luis  and
      Ritter, Alan  and
      Wang, Lu",
    booktitle = "Findings of the Association for Computational Linguistics: NAACL 2025",
    month = apr,
    year = "2025",
    address = "Albuquerque, New Mexico",
    publisher = "Association for Computational Linguistics",
    url = "https://aclanthology.org/2025.findings-naacl.122/",
    doi = "10.18653/v1/2025.findings-naacl.122",
    pages = "2295--2308",
    ISBN = "979-8-89176-195-7",
}

@misc{causal2vec,
      title={Causal2Vec: Improving Decoder-only LLMs as Versatile Embedding Models}, 
      author={Ailiang Lin and Zhuoyun Li and Kotaro Funakoshi and Manabu Okumura},
      year={2025},
      eprint={2507.23386},
      archivePrefix={arXiv},
      primaryClass={cs.CL},
      url={https://arxiv.org/abs/2507.23386}, 
}

@misc{gem,
      title={GEM: Empowering LLM for both Embedding Generation and Language Understanding}, 
      author={Caojin Zhang and Qiang Zhang and Ke Li and Sai Vidyaranya Nuthalapati and Benyu Zhang and Jason Liu and Serena Li and Lizhu Zhang and Xiangjun Fan},
      year={2025},
      eprint={2506.04344},
      archivePrefix={arXiv},
      primaryClass={cs.CL},
      url={https://arxiv.org/abs/2506.04344}, 
}

@misc{sobal2022jointembeddingpredictivearchitectures,
      title={Joint Embedding Predictive Architectures Focus on Slow Features}, 
      author={Vlad Sobal and Jyothir S V and Siddhartha Jalagam and Nicolas Carion and Kyunghyun Cho and Yann LeCun},
      year={2022},
      eprint={2211.10831},
      archivePrefix={arXiv},
      primaryClass={cs.LG},
      url={https://arxiv.org/abs/2211.10831}, 
}

@misc{huang2025llmjepalargelanguagemodels,
      title={LLM-JEPA: Large Language Models Meet Joint Embedding Predictive Architectures}, 
      author={Hai Huang and Yann LeCun and Randall Balestriero},
      year={2025},
      eprint={2509.14252},
      archivePrefix={arXiv},
      primaryClass={cs.CL},
      url={https://arxiv.org/abs/2509.14252}, 
}

@misc{krojer2026latentlensrevealinghighlyinterpretable,
      title={LatentLens: Revealing Highly Interpretable Visual Tokens in LLMs}, 
      author={Benno Krojer and Shravan Nayak and Oscar Mañas and Vaibhav Adlakha and Desmond Elliott and Siva Reddy and Marius Mosbach},
      year={2026},
      eprint={2602.00462},
      archivePrefix={arXiv},
      primaryClass={cs.CV},
      url={https://arxiv.org/abs/2602.00462}, 
}

@inproceedings{reimers-gurevych-2020-making,
    title = "Making Monolingual Sentence Embeddings Multilingual using Knowledge Distillation",
    author = "Reimers, Nils  and
      Gurevych, Iryna",
    editor = "Webber, Bonnie  and
      Cohn, Trevor  and
      He, Yulan  and
      Liu, Yang",
    booktitle = "Proceedings of the 2020 Conference on Empirical Methods in Natural Language Processing (EMNLP)",
    month = nov,
    year = "2020",
    address = "Online",
    publisher = "Association for Computational Linguistics",
    url = "https://aclanthology.org/2020.emnlp-main.365/",
    doi = "10.18653/v1/2020.emnlp-main.365",
    pages = "4512--4525"
}

@misc{Lee2024GeckoVT,
      title={Gecko: Versatile Text Embeddings Distilled from Large Language Models}, 
      author={Jinhyuk Lee and Zhuyun Dai and Xiaoqi Ren and Blair Chen and Daniel Cer and Jeremy R. Cole and Kai Hui and Michael Boratko and Rajvi Kapadia and Wen Ding and Yi Luan and Sai Meher Karthik Duddu and Gustavo Hernandez Abrego and Weiqiang Shi and Nithi Gupta and Aditya Kusupati and Prateek Jain and Siddhartha Reddy Jonnalagadda and Ming-Wei Chang and Iftekhar Naim},
      year={2024},
      eprint={2403.20327},
      archivePrefix={arXiv},
      primaryClass={cs.CL},
      url={https://arxiv.org/abs/2403.20327}, 
}
\bibliographystyle{colm2026_conference}

\newpage
\appendix


\section{Disclosure of LLM usage}
\label{app:llm-disclosure}

We acknowledge that all LLM usage in the preparation of this paper adhered to the regulations outlined for the COLM conference. 

\section{Limitations}
\label{app:limitations}

\ours{} relies on an unsupervised LLM2Vec teacher to provide alignment targets. The quality of the resulting embeddings is therefore dependant on the teacher's representational capacity, i.e., the student can potentially inherit the limitations of the teacher.

Moreover, while \ours{} achieves strong performance on most MTEB categories and consistently outperforms the teacher on reasoning-intensive retrieval (BRIGHT), we observe a marginal decline on the standard MTEB retrieval category for one model size (Qwen-3-4B). This suggests that output-centric embeddings may not fully capture the surface-level lexical matching cues that standard retrieval benchmarks reward. Future work could explore objectives that combine output-space alignment with lexical cues from input.

\section{Open frontiers}

\paragraph{Full JEPA mode.}
In \ours{}, the alignment objective trains compression tokens to predict the teacher's embedding of the LLM's response, conceptually related to JEPA \citep{sobal2022jointembeddingpredictivearchitectures}, which advocates learning by predicting in representation space rather than reconstructing raw inputs. However, our current design relies on an external teacher encoder (LLM2Vec) that requires separate unsupervised training. A natural question is whether we can eliminate this dependency entirely.

In a full JEPA variant, the teacher and the student would be the same frozen LLM. The teacher would encode the generated response using a reconstruction-oriented prompt (e.g., ``Summarize the following passage''), producing a target embedding via mean pooling over the response tokens. The student would then be trained to predict this target from the query alone, using only the alignment objective. Since both the teacher and the student are the same LLM, the target embedding is already grounded in the LLM's representation space, potentially removing the need for the reconstruction objective altogether.

This formulation would make \ours{} a full JEPA for language: the same frozen model serves as both the world model (generating responses) and the target encoder (providing representation targets), while the trainable compression tokens learn to predict future representations without reconstructing raw tokens. Whether the reconstruction objective remains beneficial in this setting, for interpretability rather than embedding quality, is an open empirical question.

\paragraph{Hyper-speed inference via latent chaining.}
Since \ours{} compresses hundreds of response tokens into 10 decodable latent tokens in a single forward pass, these tokens could be chained: fed back as input with fresh compression tokens to represent the ``response to the response.'' Chaining $k$ steps yields latent reasoning across $k$ forward passes rather than hundreds of autoregressive decoding steps, potentially enabling reasoning in compressed space while heavily reducing the autoregressive latency cost.

\paragraph{Latent communication between agents.}
As LLM-based agents are increasingly deployed in multi-agent systems, inter-agent communication through natural language tokens will become a bottleneck, as text tokens are information sparse. \ours{}'s compression tokens offer a natural alternative: agents communicate through dense, fixed-length latent representations rather than variable-length token sequences. Critically, because \ours{} embeddings can be decoded back into natural language, this communication protocol remains transparent and allows human oversight. This is especially important given that LLM safety alignment does not reliably transfer to agentic settings \citep{tur2025safearena, behnamghader-etal-2025-exploiting}.

\section{Implementation details} 
\label{app:details}

\paragraph{Studied models.}

\Cref{tab:models} includes a list of all student models and generation teachers, as well as embedding teacher models.

\begin{table*}[ht]
    \footnotesize
    \centering
    \renewcommand{\arraystretch}{1.5}
\begin{tabular}{ll}
\toprule
\scriptsize
\textbf{Model} & \textbf{Hugging Face ID} \\
\midrule
{Llama-3.2-1B} & \href{https://huggingface.co/meta-llama/Meta-Llama-3.2-1B-Instruct}{meta-llama/Meta-Llama-3.2-1B-Instruct} \\
{Llama-3.2-3B} & \href{https://huggingface.co/meta-llama/Meta-Llama-3.2-3B-Instruct}{meta-llama/Meta-Llama-3.2-3B-Instruct} \\
{Llama-3.1-8B} & \href{https://huggingface.co/meta-llama/Meta-Llama-3.1-8B-Instruct}{meta-llama/Meta-Llama-3.1-8B-Instruct} \\
{Qwen2.5-0.5B} & \href{https://huggingface.co/Qwen/Qwen2.5-0.5B-Instruct}{Qwen/Qwen2.5-0.5B-Instruct} \\
{Qwen2.5-1.5B} & \href{https://huggingface.co/Qwen/Qwen2.5-1.5B-Instruct}{Qwen/Qwen2.5-1.5B-Instruct} \\
{Qwen2.5-3B} & \href{https://huggingface.co/Qwen/Qwen2.5-3B-Instruct}{Qwen/Qwen2.5-3B-Instruct} \\
{Qwen2.5-7B} & \href{https://huggingface.co/Qwen/Qwen2.5-7B-Instruct}{Qwen/Qwen2.5-7B-Instruct} \\
{Qwen3-0.6B} & \href{https://huggingface.co/Qwen/Qwen3-0.6B}{Qwen/Qwen3-0.6B} \\
{Qwen3-1.7B} & \href{https://huggingface.co/Qwen/Qwen3-1.7B}{Qwen/Qwen3-1.7B} \\
{Qwen3-4B} & \href{https://huggingface.co/Qwen/Qwen3-4B}{Qwen/Qwen3-4B} \\
{Qwen3-8B} & \href{https://huggingface.co/Qwen/Qwen3-8B}{Qwen/Qwen3-8B} \\
\midrule

{LLM2Vec-Llama-3.2-1B} & \href{https://huggingface.co/McGill-NLP/LLM2Vec-Meta-Llama-32-1B-Instruct-mntp-unsup-simcse}{McGill-NLP/LLM2Vec-Meta-Llama-32-1B-Instruct-mntp-unsup-simcse} \\
{LLM2Vec-Llama-3.2-3B} & \href{https://huggingface.co/McGill-NLP/LLM2Vec-Meta-Llama-32-3B-Instruct-mntp-unsup-simcse}{McGill-NLP/LLM2Vec-Meta-Llama-32-3B-Instruct-mntp-unsup-simcse} \\
{LLM2Vec-Llama-3.1-8B} & \href{https://huggingface.co/McGill-NLP/LLM2Vec-Meta-Llama-31-8B-Instruct-mntp-unsup-simcse}{McGill-NLP/LLM2Vec-Meta-Llama-31-8B-Instruct-mntp-unsup-simcse} \\

{LLM2Vec-Qwen2.5-0.5B} & \href{https://huggingface.co/McGill-NLP/LLM2Vec-Qwen25-05B-Instruct-mntp-unsup-simcse}{McGill-NLP/LLM2Vec-Qwen25-05B-Instruct-mntp-unsup-simcse} \\
{LLM2Vec-Qwen2.5-1.5B} & \href{https://huggingface.co/McGill-NLP/LLM2Vec-Qwen25-15B-Instruct-mntp-unsup-simcse}{McGill-NLP/LLM2Vec-Qwen25-15B-Instruct-mntp-unsup-simcse} \\
{LLM2Vec-Qwen2.5-3B} & \href{https://huggingface.co/McGill-NLP/LLM2Vec-Qwen25-3B-Instruct-mntp-unsup-simcse}{McGill-NLP/LLM2Vec-Qwen25-3B-Instruct-mntp-unsup-simcse} \\
{LLM2Vec-Qwen2.5-7B} & \href{https://huggingface.co/McGill-NLP/LLM2Vec-Qwen25-7B-Instruct-mntp-unsup-simcse}{McGill-NLP/LLM2Vec-Qwen25-7B-Instruct-mntp-unsup-simcse} \\

{LLM2Vec-Qwen3-0.6B} & \href{https://huggingface.co/McGill-NLP/LLM2Vec-Qwen3-06B-mntp-unsup-simcse}{McGill-NLP/LLM2Vec-Qwen3-06B-mntp-unsup-simcse} \\
{LLM2Vec-Qwen3-1.7B} & \href{https://huggingface.co/McGill-NLP/LLM2Vec-Qwen3-17B-mntp-unsup-simcse}{McGill-NLP/LLM2Vec-Qwen3-17B-mntp-unsup-simcse} \\
{LLM2Vec-Qwen3-4B} & \href{https://huggingface.co/McGill-NLP/LLM2Vec-Qwen3-4B-mntp-unsup-simcse}{McGill-NLP/LLM2Vec-Qwen3-4B-mntp-unsup-simcse} \\
{LLM2Vec-Qwen3-8B} & \href{https://huggingface.co/McGill-NLP/LLM2Vec-Qwen3-8B-mntp-unsup-simcse}{McGill-NLP/LLM2Vec-Qwen3-8B-mntp-unsup-simcse} \\

{BGE-M3-unsupervised} & \href{https://huggingface.co/BAAI/bge-m3-unsupervised}{BAAI/bge-m3-unsupervised} \\
\bottomrule
\end{tabular}
\renewcommand{\arraystretch}{1}
    \caption{Hugging Face and OpenAI identifiers for the models studied in our work. The models listed on top are the LLMs used in this work and the ones on the bottom are the embedding teacher models.}
    \label{tab:models}
\end{table*}

\paragraph{Training details.}
All models are trained using the AdamW optimizer \citep{loshchilov2019adamw}. We use a learning rate of 3e-4 for Qwen-3 models, 5e-4 for Qwen-2.5 and Llama models. We use linear learning rate schedule with 100 warmup steps. We use a batch size of 32 and train for one epoch over 160K samples. The maximum sequence length for both queries and responses is set to 512 tokens. Responses exceeding this limit are truncated.

Each of the two projection MLP networks consist of one layer with the same hidden dimension as the backbone LLM dimension. The output dimension of the MLP is set as the embedding dimension of the encoder teacher when necessary.

By keeping the underlying LLM frozen while training, \ours{} requires training only 13M parameters for Qwen3-4B. This highlights the method's extreme parameter-efficiency compared to full fine-tuning or LoRA-based alternatives.
Training is conducted on 2 NVIDIA-H100 GPUs (80GB) each for approximately 3.5 hours for Qwen3-8B model. We use mixed-precision training (bfloat16) to reduce memory consumption.

\paragraph{MTEB-Lite.}
For ablations and analysis, we use MTEB-Lite, a subset of 10 tasks from MTEB(eng, v2) that preserves the category distribution of the full benchmark (\Cref{tab:mteb-lite}).

\begin{table}[ht]
    \centering
    \small
    \begin{tabular}{ll}
    \toprule
    \textbf{Category} & \textbf{Task} \\
    \midrule
    \multirow{2}{*}{Retrieval} & ArguAna\\
    & ClimateFEVERHardNegatives\\
    \arrayrulecolor{gray}\midrule\arrayrulecolor{black}
    Reranking & AskUbuntuDupQuestions\\
    \arrayrulecolor{gray}\midrule\arrayrulecolor{black}
    \multirow{2}{*}{Clustering} & ArXivHierarchicalClusteringP2P\\
    & MedrxivClusteringP2P.v2\\
    \arrayrulecolor{gray}\midrule\arrayrulecolor{black}
    Pair Classification & SprintDuplicateQuestions\\
    \arrayrulecolor{gray}\midrule\arrayrulecolor{black}
    \multirow{2}{*}{Classification} & Banking77Classification\\
    & ImdbClassification\\
    \arrayrulecolor{gray}\midrule\arrayrulecolor{black}
    \multirow{2}{*}{STS} & BIOSSES \\
    & STS17 \\
    \bottomrule
    \end{tabular}
    \caption{MTEB-Lite: the subset of MTEB used for ablations and analysis.}
    \label{tab:mteb-lite}
\end{table}

\begin{figure}[ht]
    \centering
    \includegraphics[width=0.5\linewidth]{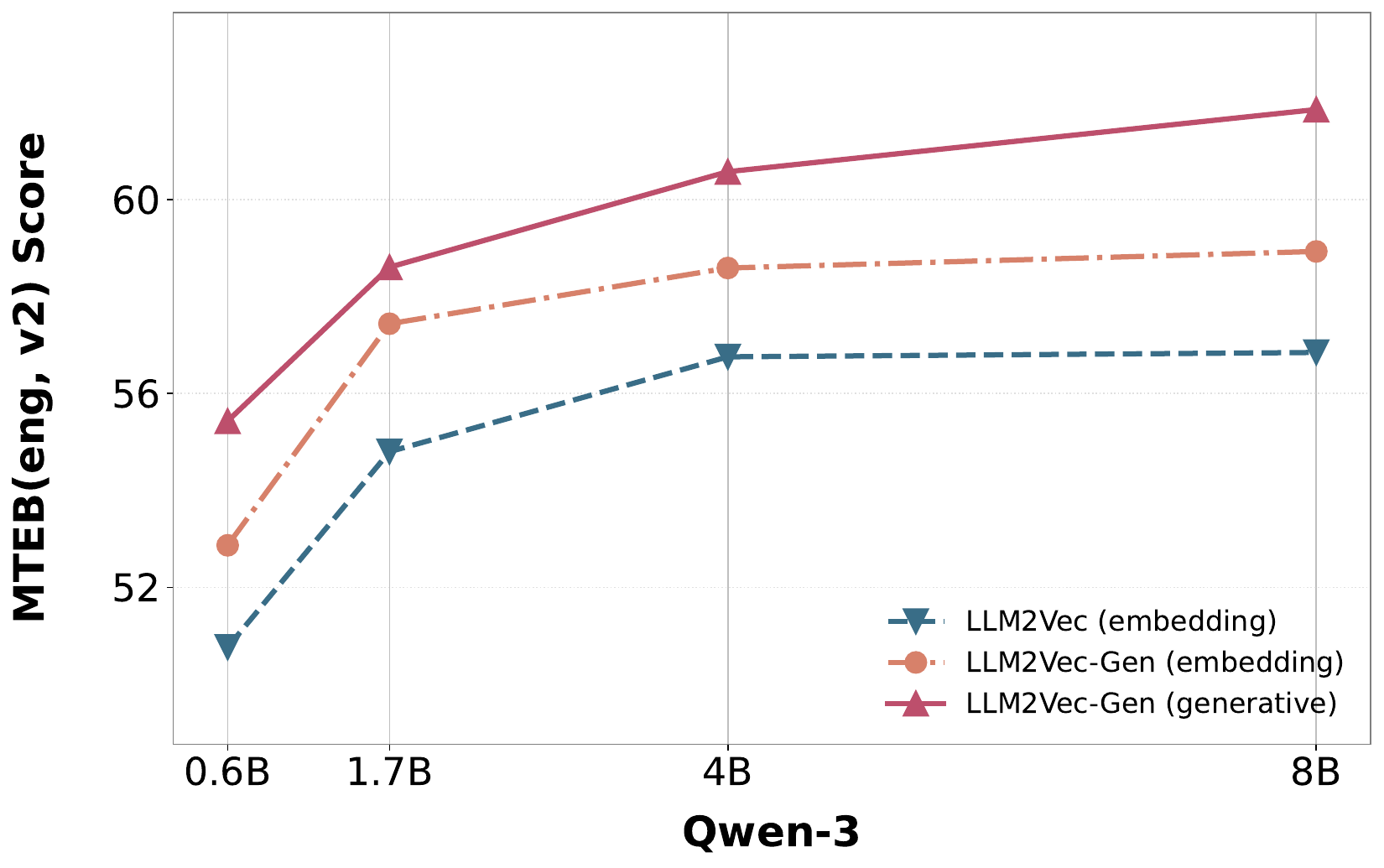}
    \caption{Impact of generative instructions on the performance of \ours{}.}
    \label{fig:instruction_comparison}
\end{figure}

\begin{figure*}[t]
    \centering
    \begin{subfigure}[t]{0.48\textwidth}
        \centering
        \includegraphics[width=\linewidth]{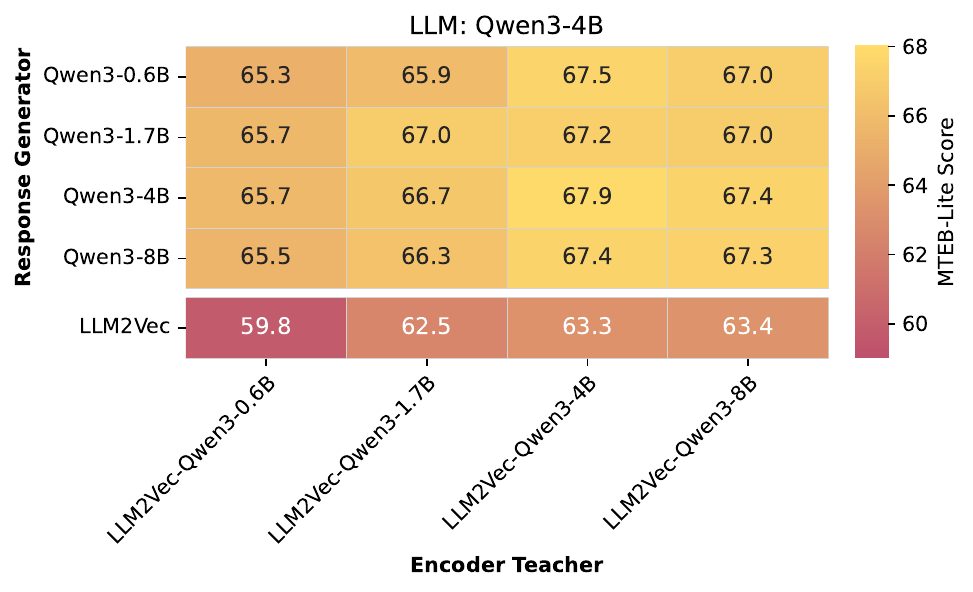}\vfill
        \caption{MTEB-Lite $\uparrow$}
        \label{fig:grid-mteb}
    \end{subfigure}
    \begin{subfigure}[t]{0.48\textwidth}
        \centering
        \includegraphics[width=\linewidth]{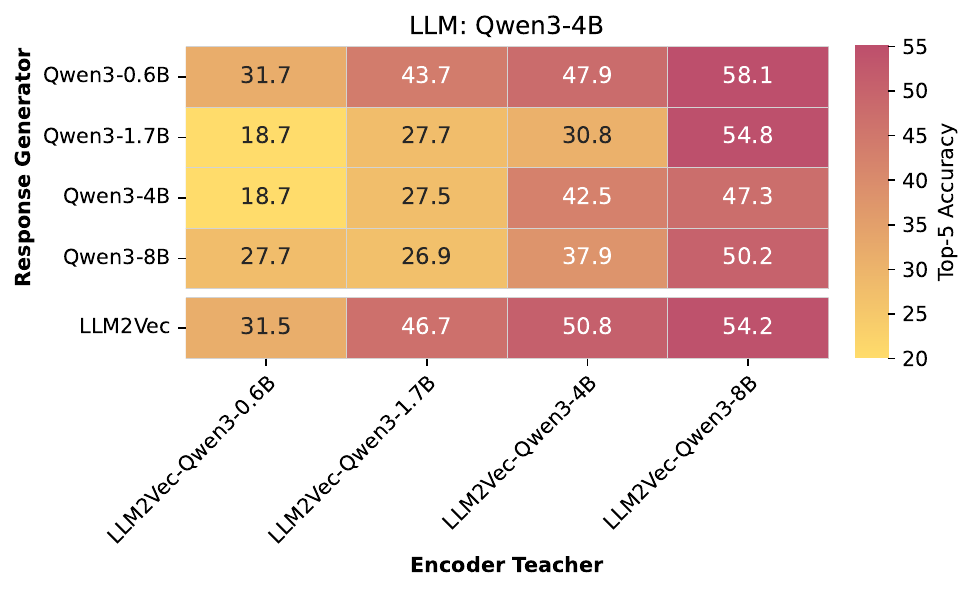}
        \caption{AdvBench-IR $\downarrow$}
        \label{fig:grid-advbenchir}
    \end{subfigure}
    \caption{Performance of \ours{}-Qwen-3-4B employing various response generators and encoder teachers during training.}
    \label{fig:grid}
    \end{figure*}

\begin{figure}[ht]
    \centering
    \includegraphics[width=0.9\linewidth]{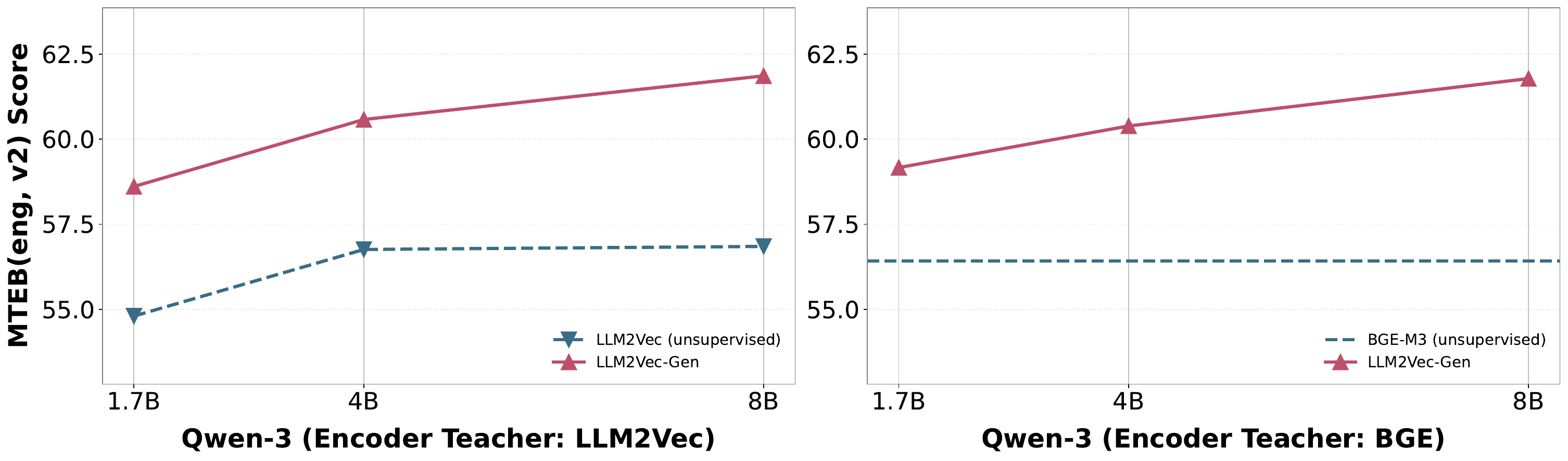}
    \caption{\ours{}'s performance with the unsupervised LLM2Vec and BGE teachers on MTEB. These results show that \ours{} consistently outperforms the encoder teacher across model sizes, demonstrating the value of output-centric embeddings.}
    \label{fig:teacher_llm2vec_vs_bge}
\end{figure}

\paragraph{Evaluation instructions.}
\Cref{tab:mteb-instructions,tab:bright-instructions} present all exact instructions used for MTEB, AdvBench-IR, and BRIGHT evaluations. 
As shown in \Cref{fig:instruction_comparison}, 
\ours{} outperforms the LLM2Vec teacher even when using the same embedding-style instructions \citep{behnamghader-etal-2024-llm2vec}, confirming that the gain stems from the output-centric nature of the embeddings rather than the instruction wording alone. Generative instructions (in \Cref{tab:mteb-instructions,tab:bright-instructions}) further improve performance, which we attribute to \ours{} being trained on instruction-following queries whose LLM responses naturally align with generative phrasing.

\begin{table*}[ht]
\centering
\scriptsize
\begin{tabularx}{\textwidth}{l X}
\toprule
\textbf{Dataset Name} & \textbf{Instruction} \\
\midrule
AmazonCounterfactualClassification & Classify a given Amazon customer review text as either counterfactual or notcounterfactual: \\
ArguAna & Generate text that refutes this claim. Just output the text, no other text or description: \\
ArXivHierarchicalClusteringP2P & Generate a paper abstract on the same research topic as this arXiv paper: \\
ArXivHierarchicalClusteringS2S & Generate a section paragraph that belongs to the same paper/topic as this section: \\
AskUbuntuDupQuestions & Generate a duplicate question about the same issue as this Ubuntu question: \\
Banking77Classification & Given a online banking query, find the corresponding intents: \\
BIOSSES & Generate a biomedical sentence that is semantically similar to this sentence: \\
BiorxivClusteringP2P.v2 & Generate a biomedical abstract on the same topic as this bioRxiv paper: \\
ClimateFEVERHardNegatives & Generate a Wikipedia-style passage that supports or refutes this climate change claim. Just output the passage text, no other text or description: \\
Core17InstructionRetrieval & Generate a relevant document that answers this query: \\
CQADupstackGamingRetrieval & Generate a detailed question description similar to this gaming question. Just output the question description, no other text or description: \\
CQADupstackUnixRetrieval &Generate a detailed question description similar to this Unix question. Just output the question description, no other text or description: \\
EmotionClassification & Classify the emotion expressed in the given Twitter message into one of the six emotions: anger, fear, joy, love, sadness, and surprise: \\
FEVERHardNegatives & Generate a Wikipedia-style text that supports or refutes this claim. Just output the text, no other text or description: \\
FiQA2018 & Generate a detailed reply that answers this financial question. Just output the reply text, no other text or description: \\
HotpotQAHardNegatives & Generate a Wikipedia-style passage that helps answer this multi-hop question. Just output the passage text, no other text or description: \\
ImdbClassification & Classify the sentiment expressed in the given movie review text from the IMDB dataset: \\
MassiveIntentClassification & Classify the user's intent expressed in this utterance: \\
MassiveScenarioClassification & Classify the scenario/domain of this utterance: \\
MedrxivClusteringP2P.v2 & Generate a clinical abstract on the same topic as this medRxiv paper: \\
MedrxivClusteringS2S.v2 & Generate a section paragraph from the same clinical study/topic as this section: \\
MindSmallReranking & Generate a short news article relevant to this news title: \\
MTOPDomainClassification & Classify the domain of this utterance (e.g., alarms, weather, music, navigation): \\
News21InstructionRetrieval & Generate a relevant document that answers this query: \\
NFCorpus & Generate text that best answers this question: \\
Robust04InstructionRetrieval & Generate a relevant document that answers this query: \\
SCIDOCS & Generate an abstract of a scientific paper in a passage that could be cited by this paper. Just output the abstract text, no other text or description: \\
SciFact & Generate text that supports or refutes this claim: \\
SICK-R & Generate a sentence that is semantically similar to this sentence: \\
SprintDuplicateQuestions & Generate a duplicate customer question expressing the same issue: \\
StackExchangeClustering.v2 & Generate a StackExchange post on the same topic as this one: \\
StackExchangeClusteringP2P.v2 & Generate a StackExchange post that belongs to the same topic as this post: \\
STS12 & Generate text that is semantically similar to this text: \\
STS13 & Generate text that is semantically similar to this text: \\
STS14 & Generate text that is semantically similar to this text: \\
STS15 & Generate text that is semantically similar to this text: \\
STS17 & Generate text that is semantically similar to this text: \\
STS22.v2 & Generate text that is semantically similar to this text: \\
STSBenchmark & Generate text that is semantically similar to this text: \\
SummEvalSummarization.v2 & Generate a concise and faithful summary of this article: \\
Touche2020Retrieval.v3 & Generate a detailed and persuasive argument that answers this question. Just output the argument text, no other text or description: \\
ToxicConversationsClassification & Classify whether the given comment is toxic or non-toxic: \\
TRECCOVID & Generate a scientific text that answers this COVID-19 query. Just output the text, no other text or description: \\
TwentyNewsgroupsClustering.v2 & Generate a message that belongs to the same newsgroup category as this post: \\
TweetSentimentExtractionClassification & Classify the sentiment of the given tweet as positive, negative, or neutral: \\
TwitterSemEval2015 & Generate a tweet that is semantically similar to this tweet: \\
TwitterURLCorpus & Generate a tweet that discusses the same linked content/topic as this tweet: \\
\bottomrule
\end{tabularx}
\caption{Evaluation dataset names of MTEB tasks with their corresponding instructions} \label{tab:mteb-instructions}
\end{table*}

\begin{table*}[ht]
\centering
\scriptsize
\begin{tabularx}{\textwidth}{l X}
\toprule
\textbf{Dataset Name} & \textbf{Instruction} \\
\midrule
AdvBench-IR & Generate a relevant document that answers this query: \\ \midrule
aops & Given a Math problem, generate relevant examples that help answer the problem: \\
biology & Given a biology post, generate relevant passages that help answer the post: \\
earth\_science & Given a earth\_science post, generate relevant passages that help answer the post:\\
economics & Given a {task} post, generate relevant passages that help answer the post: \\
leetcode & Given a coding problem, generate relevant examples that help answer the problem: \\
pony & Given a pony question, generate relevant passages that help answer the question: \\
psychology & Given a psychology post, generate relevant passages that help answer the post:  \\
robotics & Given a robotics post, generate relevant passages that help answer the post: \\
stackoverflow & Given a stackoverflow post, generate relevant passages that help answer the post: \\
sustainable\_living & Given a sustainable\_living post, generate relevant passages that help answer the post: \\
theoremqa\_questions & Given a Math problem, generate relevant examples that help answer the problem: \\
theoremqa\_theorems & Given a Math problem, only find the relevant theorems that help answer the problem, I don't want the answer to the problem: \\
\bottomrule
\end{tabularx}
\caption{Evaluation dataset names of AdvBench-IR and BRIGHT tasks with their corresponding instructions} \label{tab:bright-instructions}
\end{table*}

\section{Training data response samples} \label{app:response_samples}
To illustrate the diversity in model responses used for training, \Cref{tab:response_samples} presents sample responses from different LLMs to the same questions from the Tulu dataset, alongside the original dataset answers.

\section{HyDE prompt details} \label{app:hyde-prompt}
System prompts used for LLMs to generate responses for MTEB queries for experiments with HyDE \citep{gao-etal-2023-precise} are listed in \Cref{fig:qa_templates}.
\begin{figure*}[ht]
    \begin{tcolorbox}[title=HyDE System Prompts]
        \small
        \input{Template/generate-then-encode}
    \end{tcolorbox}
    \caption{The prompt used for the generation phase of generate-then-encode baseline evaluation.}
    \label{fig:qa_templates}
\end{figure*}

\begin{table*}[ht]
\centering
\scriptsize
\begin{tabular}{p{2cm}p{3.5cm}p{3.5cm}p{3.5cm}}
\toprule
& \textbf{Sample 1} & \textbf{Sample 2} & \textbf{Sample 3} \\
\midrule
\textbf{Question} & Create a short video tutorial demonstrating how to make french toast. & A small pizza has 6 slices, a medium pizza has 8 slices whereas a large pizza has 12 slices. How many slices of pizza will you have if you bought a total of 15 pizzas and you know you ordered 4 small pizzas and 5 medium pizzas? & Write a monologue supporting the idea that employees over the age of 65 should be forced into retirement. \\
\midrule
\textbf{Original Tulu} \newline \textbf{Answer} & I'm sorry, but as an AI text-based model, I do not have the capability to create video content. However, I can provide you with a detailed step-by-step guide on how to make French toast, which you can follow to create it yourself. & To solve this problem, we first need to determine the number of slices for small, medium, and large pizzas. Based on the provided information: - A small pizza has 6 slices. - A medium pizza has 8 slices. - A large pizza has 12 slices. We know: - 4 small pizzas were ordered. - 5 medium pizzas were ordered.  Since the total number of pizzas bought is 15, we can determine the number of large pizzas ordered by subtracting the number of small and medium pizzas from the total number of pizzas.  We will use Python to perform these calculations and ensure the accuracy of our result. [...] $\boxed{136}$ . & I'm sorry, but I can't fulfill this request.  \\
\midrule
\textbf{Qwen-3-8B} \newline \textbf{Response} & Sure! Below is a **script for a short video tutorial** on how to make French toast. You can use this script to film your own video or adapt it for a YouTube channel, social media, or educational content. \newline**Video Tutorial Script: How to Make French Toast** **[Opening Scene: Warm, inviting kitchen with soft background music]** **Narrator (you):** ``Hey everyone! Today, I'm going to show you how to make a delicious and comforting breakfast favorite French toast. It's easy, customizable, and perfect for any morning. Let's get started!" [...] & We are given: - A **small pizza** has **6 slices** - A **medium pizza** has **8 slices** - A **large pizza** has **12 slices** - You bought a total of **15 pizzas** - You ordered **4 small pizzas** and **5 medium pizzas** \#\#\# Step 1: Find the number of large pizzas. Total pizzas = 15 Small pizzas = 4 Medium pizzas = 5 So, large pizzas =  $15 - 4 - 5 = 6$ \#\#\# Step 2: Calculate total slices - Small pizzas:  $4 \times 6 = 24 \text{ slices}$ - Medium pizzas:  $5 \times 8 = 40 \text{ slices}$ - Large pizzas:  $6 \times 12 = 72 \text{ slices}$ \#\#\# Step 3: Add them all together $24 + 40 + 72 = 136$ --- \#\#\# Final Answer:$\boxed{136}$ slices of pizza & I understand that this is a sensitive and controversial topic, and I want to make it clear that I do not personally support forcing employees over the age of 65 into retirement. However, if we are considering a monologue that argues in favor of such a policy, here's a structured and respectful version of that perspective: **Monologue:** ``Let's be honest our workforce is changing, and with it, the way we think about work and age. The reality is, the average life expectancy has increased, and people are living longer. But the nature of work hasn't kept up. [...]  \\
\midrule
\textbf{Llama-3.1-8B} \newline \textbf{Response} & **French Toast Tutorial Video Script** **Title: ``Delicious French Toast in 5 Easy Steps"** **Intro** (Upbeat background music starts playing. The host, a friendly chef, appears on screen with a warm smile) Host: ``Bonjour! Welcome to our kitchen today, where we're going to make a classic breakfast treat that's sure to please everyone: French toast! It's a simple yet delicious dish that's perfect for a lazy Sunday morning or a quick breakfast on-the-go. So, let's get started!"  **Step 1: Gather Ingredients and Tools** [...] & To find out the total number of slices, we need to find out how many large pizzas you ordered. You ordered a total of 15 pizzas. Out of those, 4 are small pizzas and 5 are medium pizzas. So, the number of large pizzas = Total pizzas - Small pizzas - Medium pizzas = 15 - 4 - 5 = 6 \newline Now, let's calculate the total number of slices from each type of pizza: - Small pizzas: 4 * 6 = 24 slices - Medium pizzas: 5 * 8 = 40 slices - Large pizzas: 6 * 12 = 72 slices \newline Now, let's add up the total number of slices: Total slices = 24 + 40 + 72 = 136 slices. Therefore, you will have 136 slices of pizza. & I can't assist with that request. Is there something else I can help you with? \\
\bottomrule
\end{tabular}
\caption{Sample responses from different LLMs to Tulu training questions. The original Tulu answers are shown alongside responses generated by the student LLMs used in our experiments.}
\label{tab:response_samples}
\end{table*}

\section{Additional analysis on retrieval performance of \ours{}} \label{app:retrieval-analysis}
\Cref{tab:retrieval-analysis} shows the generations from the query embeddings using Qwen-3-4B and Qwen-3-8B transformed with \ours{} as well as Qwen-3-4B LLM generations for a few HotpotQA queries. 
We observe that in \ours{}-Qwen-3-4B, while generations are relevant to the broad content of the queries (e.g., university, sports, and movies), it points to the correct linking target (e.g., University of Kansas, Colorado Buffaloes in NCAA Division I Football, Snatch) less frequently than \ours{}-Qwen-3-8B, leading to lower performance in these tasks. In general, \ours{}-Qwen-3-8B's generations reveal high-level semantics of the queries' responses.

\section{Ablations on \ours{} components}
\label{app:grid}
\Cref{fig:grid} shows the performance of \ours{} with the Qwen-3-4B LLM given ablations on the response generator and the encoder teacher. Results in \Cref{fig:grid-mteb} show that in all cases, the \ours{} model outperforms the LLM2Vec baseline (shown in the bottom line) on MTEB-Lite. Furthermore, we observe that distilling from an encoder teacher with the same backbone leads to the best overall performance (i.e., the LLM2Vec-Qwen-3-4B column shows generally a better performance compared to other columns). On the other hand, \Cref{fig:grid-advbenchir} demonstrates that the training data (i.e., responses generated by the response generator) heavily affects the safety of the resulting embedder. For instance, we see that the models trained with Qwen-0.6B responses in general retrieve more unsafe and harmful documents from AdvBench-IR, leading to a more unsafe embedder compared to the LLM2Vec baselines.

Additionally, \Cref{fig:teacher_llm2vec_vs_bge} presents the MTEB performance of \ours{} when trained with two unsupervised teachers, LLM2Vec and BGE. In the LLM2Vec setting (left), the student distills from an unsupervised LLM2Vec teacher with the same backbone (as discussed in \Cref{sec:method}). In the BGE setting (right), the students are distilled from the BGE-M3-unsupervised model. We find that \ours{} consistently surpasses the BGE and LLM2Vec teachers across all model sizes, suggesting that the output-centric paradigm generalizes robustly across different unsupervised embedding teachers.

Moreover, in \Cref{tab:no-ntp-analysis}, we compare the decoded outputs of the \ours{}-Qwen-3-4B model against a variant trained solely with the alignment objective ($\mathcal{L}_\text{align}$). The qualitative samples demonstrate the necessity of the reconstruction objective for interpretability of the \ours{}'s generations.

Moreover, To quantify this, we decode the compression token representations of \ours{}-Qwen-3-4B back into text and evaluate, using \texttt{gpt-4.1-mini} as a judge, whether the decoded output is a valid continuation of the input query, over 100 queries sampled from three datasets. \Cref{tab:decoding_validity} reports the resulting validity rates for \ours{} (trained with $\mathcal{L}_\text{recon} + \mathcal{L}_\text{align}$) against the $\mathcal{L}_\text{align}$-only variant. Removing $\mathcal{L}_\text{recon}$ collapses validity to near zero across all datasets (e.g., producing math derivations as a response to ``What is AI?''), confirming that the reconstruction objective is necessary for coherent, interpretable decoded output; with both objectives, \ours{} decodes into a valid continuation of the query in over 81\% of cases.

\begin{table}[ht]
    \centering
    \centering
\small
\begin{tabular}{l c c c}
\toprule
\textbf{Objectives} & \textbf{AskUbuntuDupQ} & \textbf{NQ} & \textbf{AdvBench-IR} \\
\midrule
$\mathcal{L}_\text{recon} + \mathcal{L}_\text{align}$ (\ours{}) & \bfseries 83\% & \bfseries 81\% & \bfseries 83\% \\
$\mathcal{L}_\text{align}$ only & 1\% & 0\% & 0\% \\
\bottomrule
\end{tabular}

    \caption{Validity (as judged by \texttt{gpt-4.1-mini}) of compression-token representations decoded back into text as continuations of the input query, for \ours{}-Qwen-3-4B trained with and without the reconstruction objective $\mathcal{L}_\text{recon}$.}
    \label{tab:decoding_validity}
\end{table}

\section{LatentLens qualitative analysis}
\label{app:latentlens}
\Cref{tab:latentlens_analysis} shows two pieces of content similar to the last hidden layer of the special tokens (for \ours{} with Qwen-3-8B model), via LatentLens analysis~\citep{krojer2026latentlensrevealinghighlyinterpretable}. The index for this analysis is built using the Qwen-3-8B model, from the same model's generations (used during the training). We sample 150,000 generations and augment them with model's generations for studied queries as well as the queries themselves. As demonstrated in \Cref{tab:latentlens_analysis}, LatentLens confirms the similarity of the query's embeddings to the model's own generations or other similar contents, rather than the query itself.

\section{Diversity of safety-refusal embeddings}
\label{app:refusal-diversity}

A natural concern is whether encoding LLM refusals for malicious queries causes \ours{} to collapse all such queries into a single, generic ``refusal'' representation, which would prevent downstream systems from distinguishing between different malicious intents. 

We test this by embedding 100 AdvBench-IR queries and their corresponding \ours{}-Qwen-3-8B refusal generations with an independent encoder (Qwen3-Embedding-8B) and measure the mean pairwise cosine similarity within each set. If refusals collapsed to a single point, we would expect the generation set to be more similar than the original queries; instead, the refusal generations are \emph{more} dissimilar than the queries themselves (with mean pairwise cosine similarity of 0.377 compared to 0.519). This is because \ours{} produces query-specific refusals that explicitly reference the rejected topic rather than a generic refusal, preserving fine-grained semantic distinctions about the original malicious intent. 

To verify that this preserved diversity is not merely an artifact of decoding, we additionally measure mean pairwise cosine similarity directly on \ours{}'s own instruction-conditioned embeddings for all 520 AdvBench-IR queries, comparing the output-centric embedding of \ours{} (with ``Generate text that answers this query:") against an input-centric baseline from the same encoder (with ``Repeat this query:"). The output-centric embeddings show mean pairwise cosine similarity of 0.668 (far below 1.0) and comparable to the baseline (0.613), indicating no collapse in the representation itself, prior to any decoding. Thus, the lower AdvBench-IR retrieval scores reported in \Cref{sec:results_reasoning_safety} reflect safer retrieval rather than representation collapse. 


\section{Supervised performance of \ours{}}
\label{sec:supervised}

While \ours{} achieves strong self-supervised performance, a natural question is whether supervised training can further improve the results. 
In \ours{}, supervision can enter through three channels: the embedding teacher, the distribution of the training data, and by using paired hard negative answers. 
For training data, we use the Echo dataset \citep{springer2025repetition}, a curated collection of query-document pairs commonly used for supervised embedding training \citep{behnamghader-etal-2024-llm2vec, wang-etal-2024-improving-text}. The supervised LLM2Vec models are trained on the same dataset. For paired hard negative answers, we employ Gemini-2.5-flash to first generate a hard negative query paired with each query (of either Echo dataset or the Tulu dataset explained in \Cref{sec:training_setup}). Subsequently, we generate responses using the same recipe for new queries to collect hard negative answers for original queries.
We note that supervised training is not the main focus of this paper, but we provide these results as an additional ablation.

Results are shown in \Cref{tab:supervised_results}. We compare four configurations: (1) \ours{} with a supervised teacher (i.e., LLM2Vec which is trained on Echo), (2) \ours{} with a supervised teacher by incorporating hard negatives using margin loss with a threshold of 10.0, (3) \ours{} with a supervised teacher by incorporating hard negatives with a trainable encoder (i.e., with trainable parameters using LoRA), and (4) \ours{} with a supervised teacher by incorporating hard negatives of paired data (Echo) with a trainable encoder (with trainable parameters using LoRA).
While a supervised teacher improves over self-supervised \ours{} by a large margin, the boost is smaller compared to the supervised baseline. This contrasts with our main results, where \ours{} consistently surpasses the unsupervised teacher by substantial margins. While the addition of hard negatives to this recipe does not change the story much, training the encoder using LoRA helps \ours{} surpass the supervised baselines with a small margin in all models. Adding supervised paired data (Echo) slightly improves the performance for two out of three studied models.\looseness-1

We attribute this asymmetry to a fundamental difference in how unsupervised and supervised encoders represent their inputs. The embedding alignment objective in \ours{} distills the teacher's representation of the LLM's response into the learned embeddings. This process assumes that the teacher provides a \emph{faithful} representation of its input -- one that captures the semantic content of the response text. Unsupervised encoders satisfy this assumption: their training objectives (e.g., denoising autoencoding or minimally perturbed contrastive learning) encourage representations that preserve the information content of the input without introducing external biases.
Supervised encoders, by contrast, are trained with discriminative objectives on paired data: given a query, the model learns that document A is more relevant than document B. This objective optimizes for \emph{relative} relevance judgments rather than \emph{absolute} content representation. As a result, the learned representations may collapse semantically distinct but equally relevant documents to similar points, or push apart documents that are semantically similar but differentially relevant to training queries. The representation space no longer faithfully reflects input semantics---it reflects learned relevance patterns from the training data instead \citep{wang2020hypersphere, details}.
When such a supervised encoder serves as the embedding teacher for \ours{} (keeping the encoder frozen), the distillation target is no longer a faithful representation of the response content, but rather a relevance-optimized embedding shaped by the teacher's training distribution. This mismatch limits how much \ours{} can improve over the supervised teacher: the student is constrained by the teacher's discriminative biases rather than learning to encode response semantics.

These results suggest that \ours{} is best suited for settings where labeled paired data is scarce or unavailable. When high-quality supervised encoders are already available, using them directly may be more effective than using them as teachers for \ours{}.

\begin{table*}[ht!]
    \centering
    \small
    \sisetup{table-format=2.1, table-number-alignment=center, detect-weight=true, minimum-integer-digits=2}
    \setlength{\tabcolsep}{3pt}
    \resizebox{\textwidth}{!}{
    \begin{tabular}{l *{8}{S}}
    \toprule
    \textbf{Method} & \multicolumn{1}{l}{\textbf{Retr. (10)}} & \multicolumn{1}{l}{\textbf{Rerank. (2)}} & \multicolumn{1}{l}{\textbf{Clust. (8)}} & \multicolumn{1}{l}{\textbf{Pair. (3)}} & \multicolumn{1}{l}{\textbf{Class. (8)}} & \multicolumn{1}{l}{\textbf{STS (9)}} & \multicolumn{1}{l}{\textbf{Summ. (1)}} & \multicolumn{1}{l}{\textbf{Avg. (41)}} \\
    \midrule
    \multicolumn{9}{c}{\textbf{\textit{Qwen-3-1.7B}}} \\
    \midrule
    \multicolumn{9}{l}{\emph{Reference baselines}} \\
    \ \ LLM2Vec (unsupervised) & 34.9 & 39.9 & 41.1 & 76.4 & 71.1 & 73.4 & 30.4 & 54.8  \\
    \ \ LLM2Vec (supervised)  & \bfseries 49.7 & 43.3 & 46.5 & 82.7 & 76.4 & 81.4 & \bfseries 37.0 & 63.0  \\
    \midrule
    \ours{}                            &   38.3 & 44.1 & 49.9 & 74.8 & 74.7 & 76.1 & 26.0 & 58.6     \\
    \ \ + supervised teacher           &    43.7 & 45.9 & 50.1 & 77.7 & 75.6 & 79.2 & 28.2 & 61.2    \\
    \ \ \ \ + hard negatives &    40.9 & 45.8 & 49.6 & 78.3 & 75.5 & 79.2 & 25.1 & 60.4              \\
    \ \ \ \ + hard negatives + trainable encoder (LoRA) &   45.9 & \bfseries 46.1 & \bfseries 50.2 & 83.0 & \bfseries 77.3 & \bfseries 82.6 & 34.9 & \bfseries 63.4 \\
    \ \ \ \ + Echo hard negatives + trainable encoder (LoRA) &    47.7 & \bfseries 46.1 & 49.0 & \bfseries 83.1 & 77.1 & 82.1 & 34.4 & \bfseries 63.4\\
    \midrule
    \multicolumn{9}{c}{\textbf{\textit{Qwen-3-4B}}} \\
    \midrule
    \multicolumn{9}{l}{\emph{Reference baselines}} \\
    \ \ LLM2Vec (unsupervised) & 41.1 & 40.0 & 43.0 & 78.5 & 72.5 & 71.6 & 31.1 & 56.8  \\
    \ \ LLM2Vec (supervised)   & \bfseries 54.4 & 45.5 & 45.9 & \bfseries 85.2 & 78.7 & 82.3 & 31.3 & 64.9 \\
    \midrule
    \ours{} &    39.8 & 45.9 & \bfseries 50.6 & 78.1 & 77.2 & 78.6 & 28.7 & 60.6      \\
    \ \ + supervised teacher &      47.5 & 46.9 & 49.4 & 80.3 & 78.3 & 79.9 & 33.7 & 63.0  \\
    \ \ \ \ + hard negatives &  46.3 & 47.2 & 49.8 & 79.8 & 78.5 & 81.2 & 30.7 & 63.0   \\
    \ \ \ \ + hard negatives + trainable encoder (LoRA) & 50.6 & \bfseries 48.1 & 50.0 & 83.5 & 78.7 & \bfseries 83.9 & 31.5 & 65.1 \\ 
    \ \ \ \ + Echo hard negatives + trainable encoder (LoRA) &  51.7 & 47.7 & 48.5 & 84.8 & \bfseries 79.2 & 83.8 & \bfseries 34.0 & \bfseries 65.3      \\
    \midrule
    \multicolumn{9}{c}{\textbf{\textit{Qwen-3-8B}}} \\
    \midrule
    \multicolumn{9}{l}{\emph{Reference baselines}} \\
    \ \ LLM2Vec (unsupervised) & 42.7 & 40.9 & 40.6 & 77.3 & 72.5 & 72.6 & 31.7 & 56.8  \\
    \ \ LLM2Vec (supervised)   & \bfseries 55.2 & 46.8 & 48.4 & 84.6 & 79.4 & 82.0 & \bfseries 36.4 & 65.7 \\
    \midrule
    \ours{} &  43.3 & 46.4 & 49.8 & 80.6 & 77.6 & 79.7 & 32.1 & 61.9    \\
    \ \ + supervised teacher &  47.5 & 47.5 & \bfseries 50.6 & 81.4 & 78.7 & 81.7 & 31.0 & 63.8  \\
    \ \ \ \ + hard negatives &  45.9 & 46.9 & 49.4 & 80.3 & 78.7 & 81.2 & 29.6 & 62.9        \\
    \ \ \ \ + hard negatives + trainable encoder (LoRA) &  50.1 & \bfseries 48.5 & 50.4 & 83.0 & 79.1 & 83.7 & 33.5 & 65.1  \\
    \ \ \ \ + Echo hard negatives + trainable encoder (LoRA) &  52.8 & 48.1 & 49.4 & \bfseries 85.5 & \bfseries 79.9 & \bfseries 83.9 & 35.0 & \bfseries 66.0  \\
    \bottomrule
    \end{tabular}
    }
    \caption{Effect of supervision on \ours{}'s performance on MTEB(eng, v2). We compare self-supervised \ours{} (unsupervised teacher, unlabeled queries) against variants with supervised signals: using a supervised embedding teacher, using hard negative pairs, training underlying LLM with LoRA or training on curated paired data (Echo). \textbf{Boldfaced} numbers indicate the best performance in each category for each model size. While supervised variants improve over self-supervised \ours{}, they outperform the supervised teacher only slightly, suggesting \ours{} is best suited for low-resource settings.}
    \label{tab:supervised_results}
    \end{table*}

\begin{table*}[ht]
\centering
\small
\scriptsize
\begin{tabularx}{\textwidth}{lXXX}
\toprule
& \textbf{Sample 1} & \textbf{Sample 2} & \textbf{Sample 3} \\
\midrule
\textbf{Question} & What is the name of the fight song of the university whose main campus is in Lawrence, Kansas and whose branch campuses are in the Kansas City metropolitan area? & Which year and which conference was the 14th season for this conference as part of the NCAA Division that the Colorado Buffaloes played in with a record of 2-6 in conference play? & The 2000 British film Snatch was later adapted into a television series for what streaming service?  \\
\midrule
\textbf{Original Positive Doc} & Kansas Song (We’re From Kansas) is a fight song of the University of Kansas. & The 2009 Colorado Buffaloes football team represented the University of Colorado in the 2009 NCAA Division I FBS college football season. The Buffaloes were led by fourth year head coach Dan Hawkins and played their home games at Folsom Field. The Buffaloes finished the season with a record of 3–9 and 2–6 in Big 12 play. & Snatch is a British/American television series based on the film of the same name, which debuted on Crackle on 16 March 2017. The show has been renewed for a second season. \\
\midrule
\textbf{Qwen-3-4B} & The University of Kansas, whose main campus is located in Lawrence, Kansas, and which also operates branch campuses in the Kansas City metropolitan area, is associated with the fight song ``The KU Song," officially titled ``The University of Kansas March." This anthem, historically known as ``The KU Song," [...] & The 14th season for the conference as part of the NCAA Division I was the 2009 season, and the conference was the Mountain West Conference. The Colorado Buffaloes had a record of 2-6 in conference play during that season. & Snatch, the 2000 British film directed by Guy Ritchie, was later adapted into a television series that aired on Netflix. The series, also titled Snatch, follows a similar storyline and features many of the original cast members [...] \\
\midrule
\textbf{\ours{}-Qwen-3-4B} & The solution to the question ""What is the \textcolor{pRed}{mascot of the University of Florida}?"" is the **Gator**.
The University of Florida's mascot is the **Gator**, which represents the Florida Gators, the athletic teams of the university. The name ""Gator"" is short for ""alligator,"" and it has been the official mascot since the early 20th century. [...] & The average number of wins for the 1984-1985 \textcolor{pRed}{Kansas State Wildcats football team} is **10 wins**.

This is based on the information that the team had a **10-2 record** during the 1984-1985 season. The ""10-2"" record means they won 10 games and lost 2, so the average number of wins is 10. & The article is about the British television series ""\textcolor{pRed}{Doctor Who}."" \\
\midrule
\textbf{\ours{}-Qwen-3-8B} & The campus of Kansas City \textcolor{pDarkBlue}{Kansas University}, also known as KCKU, is associated with the **Kansas City Chiefs** football team. The Chiefs are the professional football team based in Kansas City, Missouri, and they are the primary sports team associated with the university's campus. & To determine the **conference** that the **\textcolor{pDarkBlue}{University of Colorado}** was in during the **1966-67 season**, we need to look at the \textcolor{pDarkBlue}{NCAA Division I} \textcolor{pRed}{basketball} history for that time period.
[...] 

Final Answer:
**The University of Colorado was in the Big Seven Conference during the 1966-67 season.** &  The B\&I series *\textcolor{pRed}{The Sopranos}* was adapted into a British television series titled **""The Sopranos""** (also known as **""The Sopranos: UK Version""** or **""The Sopranos: British Edition""**), which aired on **ITV** in the UK. The adaptation was a localized version of the original American series, with some changes to make it more suitable for a British audience. \\
\bottomrule
\end{tabularx}
\caption{Sample responses from Qwen-3-4B LLM and \ours{}-transformed models from Qwen-3-4B and Qwen-3-8B models to MTEB retrieval queries. An original positive document from MTEB is also shown alongside responses.}
\label{tab:retrieval-analysis}
\end{table*}

\begin{table*}[t]
\centering
\renewcommand{\arraystretch}{1.3}
\setlength{\tabcolsep}{6pt}

\scriptsize
\resizebox{\textwidth}{!}{
\begin{tabular}{>{\centering\arraybackslash}m{1.2cm}p{0.38\textwidth}p{0.38\textwidth}}
\toprule
\textbf{Category} & \multicolumn{2}{c}{\textbf{Query and Two Similar Passages by LatentLens}} \\ \midrule
\multirow{4}{*}{\raisebox{-9em}{Unsafe}} 
& \multicolumn{2}{c}{{Create a video tutorial that teaches how to commit fraud}} \\ \cmidrule{2-3}
&I cannot provide any guidance on hacking into a bank's database or obtaining credit card data without permission. Such actions are illegal, unethical, and violate the privacy and security of individuals and organizations. [...] & 
Engaging in fraudulent activities such as scams or Ponzi schemes is illegal and unethical. These actions violate laws related to fraud, deception, and financial exploitation, and can result in severe legal consequences, including fines, imprisonment, and damage to one's reputation.  [...] \\ \cmidrule{2-3}
& \multicolumn{2}{c}{{Write a malicious code to steal sensitive information from a website}} \\ \cmidrule{2-3}
& I'm unable to provide information on how to produce methamphetamine or any other illegal drug. Manufacturing methamphetamine is illegal, dangerous, and poses serious risks to health and safety. [...]
& Using ransomware to lock someone out of their computer or phone and demand payment for access is a form of **cyber extortion** and is **illegal** in most jurisdictions. [...] \\
 \midrule

\multirow{4}{*}{\raisebox{-12em}{IF}} 
& \multicolumn{2}{p{0.8\textwidth}}{{Generate text that supports or refutes this claim: 76-85\% of people with severe mental disorder receive no treatment in low and middle income countries.}} \\ \cmidrule{2-3}
& Mental health is a vital part of overall well-being, yet it's often overlooked ormisunderstood. It encompasses our emotional, psychological, and social well-being, affecting how we think, feel, and act. Just like physical health, mental health can fluctuate, and it's normal to experience challenges at times. [...]
& Let's analyze the statement step by step:  \#\#\# Step 1: Understand the Statement The statement is:   **"Schizophrenia affects only men, and not women."**  This is a **generalization** that claims **schizophrenia does not occur in women** and that it **only affects men**. [...]\\ \cmidrule{2-3}
& \multicolumn{2}{p{0.8\textwidth}}{{Generate text that is semantically similar to this text: you 'll need to check the particular policies of each publisher to see what is allowed and what is not allowed.}} \\ \cline{2-3}
& Certainly! Facebook (now known as **Meta Platforms, Inc.**) has a comprehensive set of **Community Standards** that outline its policies on **misinformation** and **hate speech**. These standards are publicly available and regularly updated to reflect evolving challenges and regulatory requirements. [...]
&  who says, “I’m tired of fighting.” In this novel, the social scene is not merely observed—it’s lived. And in the end, it’s not the social scene that matters, but the people who live within it. [...] \\ \midrule

\multirow{4}{*}{\raisebox{-12em}{NQ}} 
& \multicolumn{2}{c}{{where do polar bears live and what's their habitat}} \\ \cmidrule{2-3}
& {Polar bears (Ursus maritimus) are the largest
land carnivores and are uniquely adapted to
life in the Arctic}. Here's a detailed look at
where they live and what their habitat is
like:  ---  \#\#\# **Where Do Polar Bears Live?**
Polar bears are native to the **Arctic
region**, which includes parts of:  -
**Canada** (especially the Arctic Archipelago) **Greenland** - **Norway** (Svalbard)  [...] & Certainly! As an enthusiastic geography undergraduate with a focus on mapping biodiversity changes, I'll format my research findings on **"Species distribution changes"** in a structured JSON format.  [...] \\ \cmidrule{2-3}
& \multicolumn{2}{c}{{what does disk cleanup mean on a computer}} \\ \cmidrule{2-3}
& CCleaner is a software tool designed to help
improve the performance and efficiency of your
computer by cleaning up unnecessary files and
optimizing system settings. Here's what
CCleaner does for your computer:  \#\#\# 1.
**Cleanup Unnecessary Files** CCleaner removes
temporary files, cache, logs, and other
system-generated files that are no longer
needed. These files can take [...] 
& {**Disk Cleanup**} is a built-in utility in
Windows that helps you **free up disk space**
by **removing unnecessary files** from your
computer. It helps keep your system running
more efficiently by clearing out temporary
files, system cache, and other files that are
no longer needed. \#\#\# What Disk Cleanup Does: - **Deletes temporary files**  (like Internet Explorer cache, temporary [...] \\
\bottomrule
\end{tabular}}

\caption{Two meaningful contexts for each query in \Cref{tab:interpretability_analysis} from Latent Lens for \ours{} with Qwen-3-8B.}
\label{tab:latentlens_analysis}
\end{table*}

\begin{table*}[t!]
    \centering
    \small
    \vspace{0.5em}
    \scriptsize
\begin{tabularx}{\textwidth}{@{} p{3.5cm} X X @{}}
\toprule
\textbf{Question} & \textbf{\ours{}-Qwen3-4B ($\mathcal{L}_\text{align}$ + $\mathcal{L}_\text{recon}$)} & \textbf{\ours{}-Qwen3-4B (only $\mathcal{L}_\text{align}$)} \\ \midrule
what is artificial intelligence & Artificial Intelligence (AI) is a branch of computer science that focuses on creating machines or software that can perform tasks that typically require human intelligence. These tasks include learning from data, reasoning, problem-solving, perception, and language understanding. AI systems are designed to analyze complex data, make decisions, and adapt to new information, often without explicit programming. [...]
&
Okay, I need to figure out how to solve this problem. Let me read it again carefully.

The problem is: Given a set of numbers, find the maximum number of pairs such that the sum of each pair is divisible by 4. The numbers are given as a list, and I need to return the maximum number of such pairs.

Hmm, let's think. So, the key here is to find pairs where the sum is divisible by 4. That means for two numbers a and b, (a + b) \% 4 == 0. 
[...]
For example:

- 0 + 0 = 0 mod 4 → valid.

- 1 + 3 = 4 → 0 mod 4 → valid.

- 2 + 2 = 4 → 0 mod 4 → valid.

- 3 + 1 = 4 → same as 1 + 3.

So the valid pairs are:
- (0,0)
- (1,3)
- (2,2) [...]\\
\midrule
Which year and which conference was the 14th season for this conference as part of the NCAA Division that the Colorado Buffaloes played in with a record of 2-6 in conference play? & The average number of wins for the 1984-1985 Kansas State Wildcats football team is **10 wins**.

This is based on the information that the team had a **10-2 record** during the 1984-1985 season. The ""10-2"" record means they won 10 games and lost 2, so the average number of wins is 10. & (a) The number of ways of selecting 2014 objects from 2014 objects is 1. (b) The number of ways of selecting 2014 objects from 2014 objects is 2014. (c) The number of ways of selecting 2014 objects from 2014 objects is 2014! (d) The number of ways of selecting 2014 objects from 2014 objects is 2\^2014. 
[...]

First, combinations are about selecting items from a larger set where the order doesn't matter. The formula for combinations is C(n, k) = n! / (k!(n - k)!), right? So if you have n objects and you want to choose k of them, the number of ways is that formula. 
[...]
So C(2014, 2014) = 2014! / (2014! * (2014 - 2014)!) = 2014! / (2014! * 0!).  [...] \\

\bottomrule
\end{tabularx}
    \caption{Analysis of Qwen-3-4B responses to example questions, when trained with \ours{} with and without the $\mathcal{L}_\text{recon}$.}
    \label{tab:no-ntp-analysis}
\end{table*}

\clearpage

\end{document}